\documentclass[11pt]{article}

\usepackage{ragged2e}   
\usepackage{pdfpages}
\usepackage[final]{acl}

\usepackage{times}
\usepackage{latexsym}

\usepackage[T1]{fontenc}

\usepackage[utf8]{inputenc}

\usepackage{microtype}

\usepackage{inconsolata}

\usepackage{graphicx}
\usepackage[most]{tcolorbox}
\tcbuselibrary{skins, breakable, listings, theorems}
\usepackage{courier} 
\usepackage{arydshln}
\definecolor{MorandiPurple}{RGB}{236,233,242}   
\definecolor{MorandiGray}{RGB}{245,245,245}     
\definecolor{MorandiGreen}{RGB}{233,243,236}    
\definecolor{MorandiLine}{RGB}{201,194,211}     

\usepackage{graphicx}
\usepackage{multirow}
\usepackage{booktabs}
\usepackage{colortbl}
\usepackage[table]{xcolor}
\usepackage{arydshln}

\usepackage[utf8]{inputenc}
\usepackage{geometry}
\usepackage{tikz}
\usepackage{booktabs}
\usepackage{array}
\usepackage{xcolor}
\usepackage{colortbl}
\usetikzlibrary{positioning}
\usepackage{amsmath}
\usepackage{enumitem}

\definecolor{primary}{RGB}{52, 152, 219}   
\definecolor{secondary}{RGB}{46, 204, 113} 
\definecolor{accent}{RGB}{155, 89, 182}    
\definecolor{background}{RGB}{236, 240, 241} 
\definecolor{textcolor}{RGB}{44, 62, 80}   
\definecolor{step1color}{RGB}{52, 152, 219} 
\definecolor{step2color}{RGB}{46, 204, 113} 
\definecolor{step3color}{RGB}{241, 196, 15} 
\definecolor{step4color}{RGB}{230, 126, 34} 
\definecolor{answercolor}{RGB}{155, 89, 182} 

\definecolor{MTitleGray}{RGB}{238,238,238}   
\definecolor{MBlockB}{RGB}{236,242,237}      
\definecolor{MBlockC}{RGB}{244,238,234}      
\definecolor{MDash}{RGB}{170,170,170}        

\newcommand{\bgB}[1]{\cellcolor{MBlockB}#1}
\newcommand{\bgC}[1]{\cellcolor{MBlockC}#1}

\usepackage{ragged2e} 

\tcbset{
  promptcard/.style={
    enhanced,
    breakable,
    arc=2mm,
    boxrule=0.6pt,
    colframe=MorandiLine,
    colback=MorandiGray,
    left=2mm,
    right=2mm,
    top=1mm,
    bottom=1mm,
    before skip=6pt,
    after skip=8pt,
    drop shadow={black!10!white},
    fontupper=\small,          
    before upper=\justifying,  
  },
  promptcardgreen/.style={
    promptcard,
    colback=MorandiGreen,
  }
}

\usepackage{booktabs, tabularx, multirow, multicol, makecell, colortbl}
\usepackage{pifont}
\usepackage{colortbl}
\usepackage{xcolor}

\definecolor{HeadMorandi}{RGB}{236,233,242}   
\definecolor{RowMorandi}{RGB}{245,245,245}    
\definecolor{SEARMorandi}{RGB}{233,243,236}   

\usepackage{enumitem}

\usepackage{bbding}

\usepackage{amsmath, amssymb}

\usepackage{marvosym}
\usepackage{float}
\usepackage{footmisc}
\usepackage{colortbl}
\usepackage{xcolor}
\definecolor{morandi}{RGB}{201, 194, 211}

\usepackage[table]{xcolor}
\definecolor{MorandiGreen}{RGB}{214,226,217} 

\usepackage[table]{xcolor}

\title{
JPO: Juris Policy Optimization for Structured Legal Reasoning  \\ in Criminal Judgment Prediction}

\newcommand{\equal}{\textsuperscript{*}}
\newcommand{\corr}{\textsuperscript{\dag}}

\author{
\textbf{Zhaolu Kang}\equal\textsuperscript{1,2},
\textbf{Yantao Liu}\equal\textsuperscript{2},\\
\textbf{Tailong Luo},
\textbf{Leqi Zheng}\textsuperscript{3},
\textbf{Lei Wei}\textsuperscript{2},
\textbf{Chenghua Zhu},\\
\textbf{Junhao Gong}\textsuperscript{2},
\textbf{Jiachen Qian}\textsuperscript{4},
\textbf{Eric Hanchen Jiang}\textsuperscript{5},
\textbf{Jiaxin Liu}\textsuperscript{6},\\
\textbf{Yuan Wang}\textsuperscript{7},
\textbf{Hao Zhang},
\textbf{Zixia Wang},
\textbf{Rong Fu},
\textbf{Zheng Lin}\textsuperscript{8},\\
\textbf{Richeng Xuan}\corr\textsuperscript{1},
\textbf{Zhichao Hu}\corr\textsuperscript{1}
\\[2pt]
\textsuperscript{1}Hunyuan, Tencent,   
\textsuperscript{2}Peking University\\
\textsuperscript{3}Tsinghua University,
\textsuperscript{4}City University of Hong Kong,
\textsuperscript{5}University of California,\\
\textsuperscript{6}University of Illinois Urbana-Champaign,
\textsuperscript{7}Zhejiang University,
\textsuperscript{8}University of Hong Kong\\
\texttt{\selectfont {zlkang25}@stu.pku.edu.cn}
}

\begin{document}
\maketitle

\begingroup
  \renewcommand\thefootnote{\fnsymbol{footnote}}
  \footnotetext[1]{Equal contribution.}
  \footnotetext[2]{Corresponding author.}
\endgroup

\begin{abstract}
Criminal judgment prediction requires models to infer statutory articles, charges, and sentencing outcomes from case facts. Unlike standard classification tasks, it involves a structured reasoning process in which statutes should be matched with facts, charges should be justified by statutes, and sentencing outcomes should remain consistent with charges. Existing approaches optimize final labels, and while some have attempted to evaluate reasoning quality, their evaluations are indirect, often relying on LLM-generated rubrics that reflect model-internal preferences rather than the inherent logical structure of legal adjudication.
We propose \textbf{Juris Policy Optimization (JPO)}, a post-training framework for structured legal reasoning in Chinese criminal judgment prediction. JPO first uses teacher-generated rationales to supervise a standardized four-step reasoning process, and then applies reinforcement learning with a composite reward over legal prediction quality, reasoning structure completeness, and cross-step consistency. JPO further introduces token-level advantage reweighting and adaptive clipping for legally salient reasoning segments.
Experiments on multiple open-source language models and three Chinese legal benchmarks show that JPO consistently improves both judgment prediction and reasoning quality over supervised fine-tuning and reinforcement learning baselines.
\end{abstract}
\section{Introduction}

Legal judgment prediction (LJP) aims to infer judicial outcomes from case facts, including applicable statutory articles, criminal charges, and sentencing decisions. As a core task in legal NLP, it has been studied to support legal information processing, case analysis, and intelligent legal assistance \citep{xiao2018cail2018, chalkidis2022lexglue, guha2023legalbenchcollaborativelybuiltbenchmark}. In criminal cases, however, judgment prediction is not merely a flat mapping from facts to labels. A plausible decision must remain coherent across several reasoning stages: the cited statutory provisions should be matched with the legally relevant facts, the predicted charge should be justified by the statutory provisions, and the final sentence should be compatible with that charge.

This structured nature is particularly important in Chinese criminal law, where judicial decisions are typically organized around factual findings, statutory articles, charges, and sentencing outcomes. As a result, high final-label accuracy alone is insufficient: a model may predict the correct charge while relying on unsupported statutes, or output a plausible sentence without a coherent legal basis.

Recent large language models (LLMs) have shown strong performance on reasoning-intensive tasks and motivated growing interest in legal-domain applications \citep{openai2023gpt, touvron2023llama, bai2023qwen, wei2022chain, kojima2023largelanguagemodelszeroshot, guha2023legalbenchcollaborativelybuiltbenchmark}. In legal judgment prediction, LLMs provide a natural way to generate both predictions and rationales. However, adapting general-purpose LLMs to criminal judgment prediction remains difficult for two reasons. First, most existing LJP datasets provide case facts and final labels, but little supervision for intermediate reasoning \citep{xiao2018cail2018, fei2024lawbench}. Models trained on such data can learn shallow correlations between facts and outcomes rather than legally grounded reasoning patterns \citep{liu2024legalduet, shi2025legalreasoner}. Second, existing post-training methods mainly optimize final outputs or general preference signals \citep{ouyang2022training, rafailov2023direct, dai2025legal, cai2025unilaw}, and while some attempt to assess reasoning quality, their evaluations are indirect, often relying on LLM-generated rubrics that reflect model-internal preferences rather than the structured dependencies that define legal adjudication \citep{lee2025evaluating, sun2025profllmbasedrewardcode}.

To address this gap, we propose \textbf{Juris Policy Optimization (JPO)}, a post-training framework for structured legal reasoning in Chinese criminal judgment prediction. JPO follows a two-stage design. In the first stage, a stronger teacher model generates rationales following a standardized four-step structure: fact extraction, statutory analysis, charge determination, and sentence prediction. These rationales are used to supervise structured reasoning generation during fine-tuning. In the second stage, we apply reinforcement learning with a composite reward that jointly captures legal prediction accuracy, reasoning structure completeness, and cross-step consistency across facts, statutes, charges, and sentences. To provide finer-grained optimization signals, JPO further introduces token-level advantage reweighting and adaptive clipping for legally salient reasoning segments.

We position JPO as a structured alignment and reward-shaping framework rather than as a resolution of the ambiguity inherent in evaluating legal reasoning. Its rewards are computable proxies. They offer more stable, interpretable, and reproducible supervision than LLM-generated rubrics, but they do not fully capture legally grounded reasoning, such as why particular facts satisfy the elements of a statute or why a neighboring charge should be excluded. Crucially, proxy does not mean arbitrary: a blind study in which three legal experts rated reasoning steps on 800 test cases shows that these proxies correlate substantially with expert judgment, with Spearman $\rho$ between 0.64 and 0.72 (Appendix~\ref{sec:appendix_proxy_expert}). They therefore track genuine reasoning quality while remaining approximations rather than a complete account of legal reasoning.

We evaluate JPO on multiple open-source LLM backbones and three Chinese legal benchmarks. Experimental results show that JPO consistently improves both judgment prediction performance and reasoning quality over supervised fine-tuning and standard reinforcement learning baselines. The gains are especially clear on compact models, suggesting that structured legal-reasoning alignment can substantially strengthen smaller open-source models for criminal judgment prediction.

Our contributions are summarized as follows:
\begin{itemize}
\vspace{-6pt}
    \item We propose \textbf{Juris Policy Optimization (JPO)}, a post-training framework for structured legal reasoning in criminal judgment prediction.
    \vspace{-8pt}
    \item We introduce a composite reinforcement learning objective that jointly optimizes legal prediction quality, reasoning completeness, and cross-step consistency.
    \vspace{-8pt}
    \item We develop token-level advantage reweighting and adaptive clipping for legally salient reasoning segments.
    \vspace{-8pt}
    \item Experiments on multiple open-source backbones and three Chinese legal benchmarks show that JPO consistently improves both judgment prediction and reasoning quality over strong baselines.
\end{itemize}

\begin{figure*}
    \centering
    \includegraphics[width=0.8\linewidth]{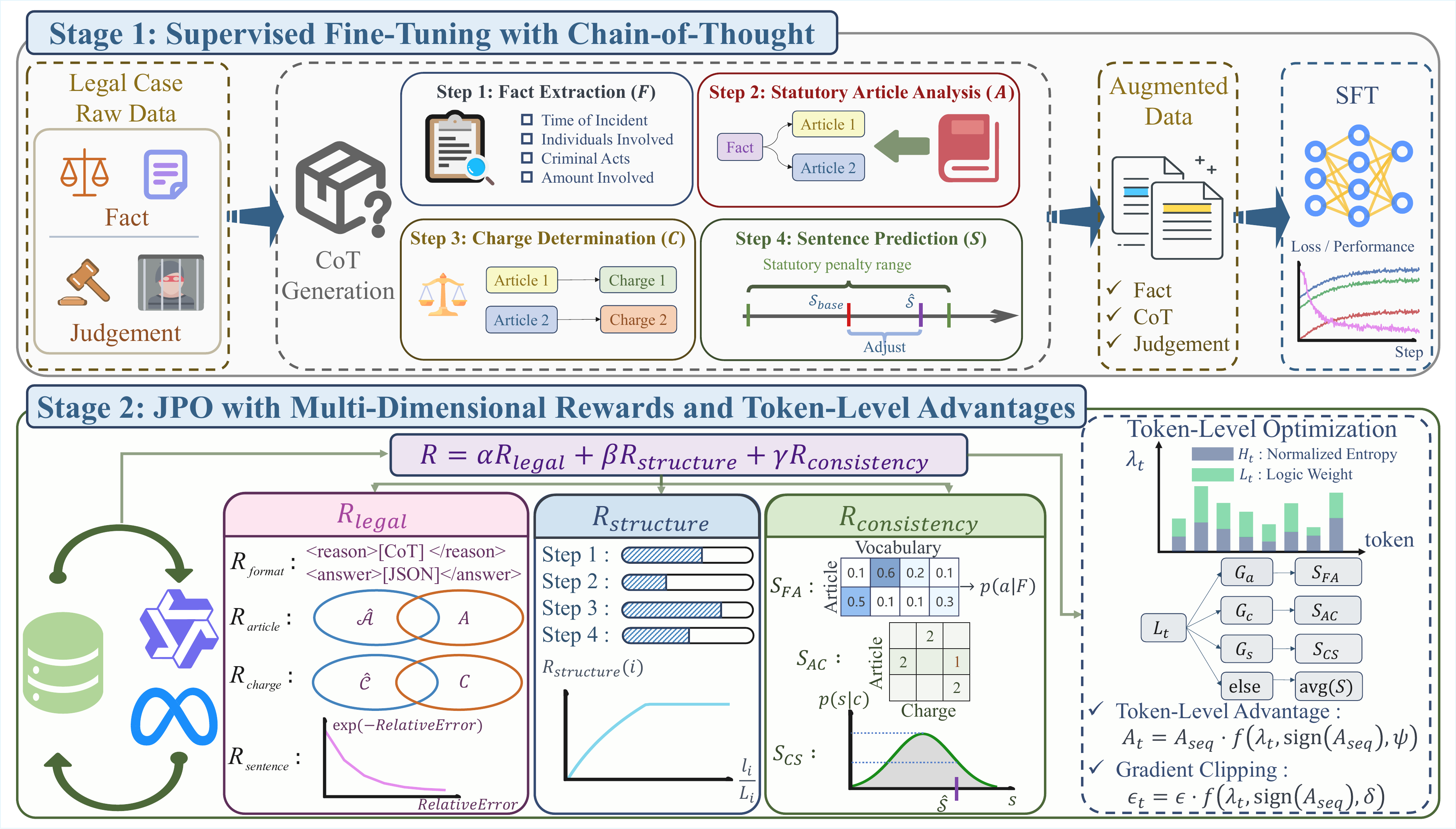}
    \vspace{-7pt}
    \caption{\textbf{Overview of JPO.} Structured SFT with teacher-generated four-step rationales, followed by reinforcement learning with rewards for legal accuracy, reasoning completeness, and cross-step consistency.}
    \vspace{-6pt}
    \label{fig:method}
\end{figure*}

\section{Related Work}

Legal judgment prediction (LJP) is a central task in legal NLP, aiming to infer judicial outcomes such as applicable statutes, charges, and sentencing results from case facts. Early work on Chinese criminal judgment prediction was largely benchmark-driven, with datasets such as CAIL establishing standard settings for article prediction, charge prediction, and prison term prediction \citep{xiao2018cail2018,kang2025jurisctc,kang2026multimodal,kang2026quanteval,he2026order,liu2026reasonact}. More recent studies have argued that accurate legal prediction requires more than final-label matching and should better capture the reasoning process connecting facts, legal rules, and judicial outcomes \citep{lee2025evaluating, liu2024legalduet, shi2025legalreasoner,yu2025trajselector,yu2026frameskip,yu-etal-2026-mathagent,yan2026sharechatdatasetchatbotconversations}. Still, most existing approaches primarily rely on supervision over final labels, without explicitly optimizing a standardized multi-stage reasoning process.

Large language models have shown strong performance on reasoning-intensive tasks and have motivated increasing interest in legal-domain adaptation and evaluation \citep{openai2023gpt, wei2022chain, guha2023legalbenchcollaborativelybuiltbenchmark, fei2024lawbench,kang2026hssbench,10.1145/3774904.3792075,feng2025seeing,gao2026laobench,shi2026spader}. In legal judgment prediction, LLMs provide a natural way to jointly generate predictions and rationales. However, adapting general-purpose LLMs to criminal judgment prediction remains difficult because most legal datasets provide limited supervision for intermediate reasoning, and unconstrained rationale generation does not by itself ensure legally grounded reasoning or cross-step coherence \cite{xie2025chat,luo2026atelierevalagenticevaluationhumans,luo2026centaurevalbenchmarkinghumanintheloopvalue,luo2026agentauditor,luo2025dynamicner,rao2026miningsynthesisrethinkingexploration,rao-etal-2026-dynamic,rao2025data}.

In parallel, post-training methods such as RLHF and DPO have become standard approaches for aligning language models with task objectives and preference signals \citep{ouyang2022training, rafailov2023direct}. Recent work has begun to extend these ideas to legal-domain reasoning through domain-specific post-training and reinforcement learning \citep{dai2025legal, cai2025unilaw}. Our work is most closely related to this line of research, but differs in a key respect: instead of optimizing only final answers or general legal preferences, we explicitly optimize the structured dependency chain from facts to statutes, charges, and sentencing outcomes in Chinese criminal judgment prediction.

An adjacent body of work combines teacher-generated reasoning traces with structured, rule-based supervision for policy and compliance reasoning, and it is worth situating JPO against it explicitly. \citet{imperial2025policyreasoningtraces} use frontier-model-generated policy reasoning traces to improve the compliance assessment and clause-citation accuracy of weaker models, which directly parallels our Stage-I use of teacher-generated four-step rationales and our statute-grounding reward. Privacy Checklist \citep{li2025privacychecklist} grounds violation detection in a regulation-derived checklist under Contextual Integrity theory and treats the task as structured reasoning rather than pattern matching, sharing the motivation behind our process-oriented metrics. Context Reasoner \citep{hu2025contextreasoner} is the closest to us methodologically, using reinforcement learning with rule-based rewards to incentivize contextualized regulatory reasoning over GDPR, the EU AI Act, and HIPAA. What separates JPO from all three is the object being supervised. These works target policy, privacy, and safety compliance, where reasoning amounts to matching a regulation against a situation. JPO instead optimizes the dependency chain of criminal adjudication, in which each step simultaneously justifies and constrains the next, and it is this chain that our cross-step consistency rewards and legal-logic-aware token reweighting are designed to exploit. Benchmark-oriented studies of Chinese legal reasoning \citep{dai2025laiw} and of judicial argumentation \citep{han2025courtreasoner} are complementary to our work, since they supply evaluation resources rather than post-training methods.

\section{Method}

We propose \textbf{Juris Policy Optimization (JPO)}, a two-stage post-training framework for structured legal reasoning in Chinese criminal judgment prediction. The central idea is to align a language model not only with the final judicial outcome, but also with the intermediate reasoning structure connecting facts, statutory articles, charges, and sentencing decisions. As illustrated in Figure~\ref{fig:method}, JPO first introduces structured supervision through teacher-generated four-step legal rationales, and then further improves the model with reinforcement learning under a composite reward that captures legal correctness, reasoning completeness, and cross-step consistency. JPO further incorporates token-level advantage reweighting and adaptive policy clipping to better allocate optimization signals within long legal responses.

\subsection{Task Definition}

Given a criminal case described by a fact sequence
\begin{equation}
F = \{f_1, f_2, \dots, f_n\},
\end{equation}
the goal of legal judgment prediction is to infer the judicial outcome
\begin{equation}
Y = (A, C, S),
\end{equation}
where $A$ denotes the applicable statutory articles, $C$ denotes the criminal charge, and $S$ denotes the sentencing outcome.

Beyond the final outcome, JPO explicitly models the reasoning trace as
\begin{equation}
Z = (G_f, G_a, G_c, G_s),
\end{equation}
where $G_f$, $G_a$, $G_c$, and $G_s$ correspond to fact extraction, statutory analysis, charge determination, and sentence prediction, respectively. These four segments define an observable reasoning template aligned with the adjudication trajectory
\begin{equation}
F \rightarrow A \rightarrow C \rightarrow S.
\end{equation}
Accordingly, JPO treats criminal judgment prediction as a structured generation problem in which the reasoning process $Z$ and the final outcome $Y$ are learned jointly.

\subsection{Stage I: Structured Supervised Fine-Tuning}

Most legal judgment prediction datasets provide case facts and final labels but do not annotate intermediate legal reasoning. JPO addresses this limitation by introducing structured reasoning supervision in the first stage.

For each training example, a stronger instruction-following teacher model generates a reasoning trace $Z$ following the four-stage structure $(G_f, G_a, G_c, G_s)$. The student model is then trained to generate both the structured trace and the final prediction conditioned on $F$:
\begin{equation}
\mathcal{L}_{\mathrm{SFT}}
=
-\log P_{\theta}(Z, Y \mid F),
\end{equation}
where $\theta$ denotes the model parameters.

This stage teaches the model a stable and interpretable reasoning template, and provides initialization for the subsequent reinforcement learning stage.

\subsection{Stage II: Juris Policy Optimization}

Although structured SFT teaches the model the desired reasoning format, it does not guarantee legally grounded reasoning. We therefore introduce a second stage based on reinforcement learning, whose reward explicitly captures both correct outcomes and coherent reasoning:
\begin{equation}
\mathcal{R}
=
\alpha R_{\mathrm{legal}}
+
\beta R_{\mathrm{structure}}
+
\gamma R_{\mathrm{consistency}},
\end{equation}
where the three components measure legal prediction quality, structural completeness, and cross-step consistency, respectively.

\subsection{Legal Prediction Reward}

The legal prediction reward evaluates whether the generated output matches the target judicial outcome. Since criminal judgment prediction contains multiple decision targets, we decompose this reward into four terms:
\begin{equation}
\begin{aligned}
R_{\mathrm{legal}}
&=
\alpha_1 R_{\mathrm{format}}
+\alpha_2 R_{\mathrm{article}} \\
&\quad
+\alpha_3 R_{\mathrm{charge}}
+\alpha_4 R_{\mathrm{sentence}}.
\end{aligned}
\end{equation}

The format reward $R_{\mathrm{format}}$ checks whether the generated response follows the required response schema, including both the reasoning trace and the final structured answer.

For article prediction, let $\hat{A}$ denote the predicted set of statutory articles. We define
\begin{equation}
R_{\mathrm{article}}
=
\frac{|\hat{A} \cap A|}{|\hat{A}|}
\cdot
\frac{|\hat{A} \cap A|}{|A|}.
\end{equation}
This reward is high only when the predicted article set is both precise and complete with respect to the gold set. In practice, when $\hat{A}$ is empty, we define $R_{\mathrm{article}}=0$.

Similarly, let $\hat{C}$ denote the predicted set of charges. The charge reward is
\begin{equation}
R_{\mathrm{charge}}
=
\frac{|\hat{C} \cap C|}{|\hat{C}|}
\cdot
\frac{|\hat{C} \cap C|}{|C|}.
\end{equation}
This formulation again encourages predictions that are simultaneously precise and complete. Likewise, when $\hat{C}$ is empty, we define $R_{\mathrm{charge}}=0$.

For sentencing prediction, exact matching is often too strict because sentencing is continuous and sensitive to case-specific aggravating and mitigating factors. We therefore define a smooth reward based on relative error:
\begin{equation}
R_{\mathrm{sentence}}
=
\exp\left(
-\xi
\frac{|\hat{S} - S|}{S}
\right),
\end{equation}
where $\hat{S}$ denotes the predicted sentence. This formulation preserves graded supervision even when the predicted sentence is only partially accurate.

\subsection{Reasoning Structure Reward}

Correct final predictions alone are insufficient if the generated reasoning is incomplete or degenerate. We therefore introduce an explicit reward for outputs that preserve the expected four-stage organization:
\begin{equation}
\begin{aligned}
R_{\mathrm{structure}}
&=
\sum_{k \in \{f,a,c,s\}}
w_k \,\mathbb{I}(G_k)\, \Bigl[ \\
&\qquad \min(\log(1+\eta \ell_k), 1)
\Bigr].
\end{aligned}
\end{equation}
where $\mathbb{I}(G_k)$ indicates whether the corresponding reasoning segment appears in the generated trace.

The normalized segment length is defined as
\begin{equation}
\ell_k
=
\frac{l_k}{L_k},
\end{equation}
where $l_k$ is the actual length of the generated segment and $L_k$ is the target minimum length for that segment.

This reward plays two complementary roles: the indicator term encourages the model to include all required reasoning stages, while the length term discourages empty or trivial segment generation.

\subsection{Logical Consistency Reward}

A response may be structurally complete yet still remain legally incoherent across reasoning steps. We therefore define a consistency reward over the three local transitions in the legal reasoning chain:
\begin{equation}
R_{\mathrm{consistency}}
=
\frac{1}{3}
\left(
S_{FA}
+
S_{AC}
+
S_{CS}
\right).
\end{equation}
Here, $S_{FA}$, $S_{AC}$, and $S_{CS}$ denote fact-to-article, article-to-charge, and charge-to-sentence consistency, respectively. These terms are designed as lightweight and computable proxies for the three adjacent reasoning transitions.

\paragraph{Fact-to-article consistency.}
We estimate whether the predicted statutory articles are supported by the case facts using a Naive Bayes article predictor trained on the corpus. Let $\mathcal{A}$ denote the full article space, and let $P(a \mid F)$ denote the posterior probability assigned to article $a \in \mathcal{A}$ for case facts $F$:
\begin{equation}
P(a \mid F)
=
\frac{
P(a)\prod_{f \in F}P(f \mid a)
}{
\sum_{a' \in \mathcal{A}}
P(a')\prod_{f \in F}P(f \mid a')
}.
\end{equation}

Given the predicted article set $\hat{A}$, we define the fact-to-article consistency score as
\begin{equation}
S_{FA}
=
\frac{1}{|\hat{A}|}
\sum_{a \in \hat{A}}
P(a \mid F).
\end{equation}
When $\hat{A}$ is empty, we define $S_{FA}=0$.

This score is high when the predicted articles are well supported by the observed case facts.

\paragraph{Article-to-charge consistency.}
We next evaluate whether the predicted charge is compatible with the predicted statutory articles through an article--charge association matrix
\begin{equation}
D \in \mathbb{R}^{|\mathcal{A}| \times |\mathcal{C}|},
\end{equation}
where $\mathcal{A}$ and $\mathcal{C}$ denote the full article space and charge space, respectively, and $D_{a,c}$ measures the strength of association between article $a$ and charge $c$.

In our implementation, $D_{a,c}$ is defined by a lightweight rule-based mapping derived from statutory references and charge annotations in the training data (see Appendix~\ref{sec:appendix_consistency_impl} for detailed provenance): $D_{a,c}=2$ if article $a$ directly governs charge $c$, $D_{a,c}=1$ if the relation is relevant but indirect, and $D_{a,c}=0$ otherwise.

Given the predicted article set $\hat{A}$ and predicted charge set $\hat{C}$, the corresponding consistency score is
\begin{equation}
S_{AC}
=
\frac{
\sum_{a \in \hat{A}}\max_{c \in \hat{C}} D_{a,c}
}{
|\hat{A}|\max(D)
}.
\end{equation}
When $\hat{A}$ or $\hat{C}$ is empty, we define $S_{AC}=0$.

This term favors reasoning traces in which the predicted charge is supported by the cited statutory basis.

\paragraph{Charge-to-sentence consistency.}
Finally, we evaluate whether the predicted sentence is plausible under the predicted charge. For each charge $c$, we estimate a charge-conditioned sentence distribution from the training data using a truncated normal distribution:
\begin{equation}
p(s \mid c)
=
\frac{
\phi\!\left(\frac{s-\mu_c}{\sigma_c}\right)
}{
\sigma_c
\left[
\Phi\!\left(\frac{b_c-\mu_c}{\sigma_c}\right)
-
\Phi\!\left(\frac{a_c-\mu_c}{\sigma_c}\right)
\right]
},
\end{equation}
where $[a_c,b_c]$ is the observed valid sentencing range for charge $c$ in the training corpus, and $\mu_c$ and $\sigma_c$ are the empirical mean and standard deviation of the sentence distribution for that charge.

The resulting consistency score is
\begin{equation}
S_{CS}
=
\frac{1}{|\hat{C}|}
\sum_{c \in \hat{C}}
\frac{
p(\hat{S} \mid c)
}{
\max_{s} p(s \mid c)
}.
\end{equation}
When $\hat{C}$ is empty, we define $S_{CS}=0$.

This term measures whether the predicted sentence lies in a plausible region of the charge-conditioned distribution.

\subsection{Token-Level Advantage Reweighting}

The rewards above are computed at the sequence level, but different tokens contribute unequally to legal reasoning quality. We therefore introduce token-level advantage reweighting. Let $A_{\mathrm{seq}}$ denote the sequence-level advantage. For token $t$,
\begin{equation}
A_t
=
A_{\mathrm{seq}}
\cdot
f(\lambda_t,\operatorname{sign}(A_{\mathrm{seq}}),\psi),
\end{equation}
where $\lambda_t \in [0,1]$ denotes the importance weight of token $t$, and $\psi$ controls the reweighting strength.

The importance score combines two complementary signals:
\begin{equation}
\lambda_t
=
\zeta H_t
+
(1-\zeta)L_t,
\end{equation}
where $H_t$ denotes the normalized entropy weight and $L_t$ denotes the legal logic weight.

\paragraph{Entropy weight.}
The entropy term measures model uncertainty at token $t$. Let $p_t(v)$ denote the predicted probability of vocabulary item $v$ at decoding step $t$. The token entropy is
\begin{equation}
\mathcal{H}_t
=
-
\sum_{v \in \mathcal{V}}
p_t(v)\log p_t(v),
\end{equation}
where $\mathcal{V}$ is the vocabulary.

Since entropy depends on vocabulary size and decoding dynamics, we normalize it as
\begin{equation}
H_t
=
\frac{\mathcal{H}_t}{\max_{\tau=1}^{T}\mathcal{H}_\tau}.
\end{equation}
A larger $H_t$ indicates higher uncertainty and thus a potentially stronger need for learning signal.

\paragraph{Legal logic weight.}
The legal logic weight depends on the reasoning segment to which token $t$ belongs. Tokens are assigned to segments according to their position in the generated reasoning trace $Z=(G_f,G_a,G_c,G_s)$. We define
\begingroup
\renewcommand{\arraystretch}{0.92}
\begin{equation}
L_t =
\begin{cases}
S_{FA}, & t \in G_a \\
S_{AC}, & t \in G_c \\
S_{CS}, & t \in G_s \\
\frac{1}{3}(S_{FA}+S_{AC}+S_{CS}), & \text{otherwise.}
\end{cases}
\end{equation}
\endgroup

Given $\lambda_t$, the reweighting function is
\begin{equation}
\begin{aligned}
&f(\lambda_t,\operatorname{sign}(A_{\mathrm{seq}}),\psi) \\
&\quad =
\begin{cases}
1+\psi(\lambda_t-\bar{\lambda}), & A_{\mathrm{seq}}>0 \\
1-\psi(\lambda_t-\bar{\lambda}), & A_{\mathrm{seq}}<0 \\
1, & A_{\mathrm{seq}}=0
\end{cases}
\end{aligned}
\end{equation}
where
\begin{equation}
\bar{\lambda}
=
\frac{1}{T}\sum_{\tau=1}^{T}\lambda_\tau
\end{equation}
is the average token importance of the sequence and $T$ is the sequence length.

When $A_{\mathrm{seq}}$ is positive, above-average important tokens receive larger positive updates. When $A_{\mathrm{seq}}$ is negative, the same tokens are penalized less aggressively, which helps preserve legally salient content during policy optimization.

\subsection{Adaptive Policy Clipping}

Standard clipped policy optimization uses a uniform clipping threshold for all tokens. We instead introduce token-aware clipping:
\begin{equation}
\epsilon_t
=
\epsilon
\cdot
f(\lambda_t,\operatorname{sign}(A_{\mathrm{seq}}),\delta),
\end{equation}
where $\epsilon$ is the base clipping coefficient and $\delta$ controls the adaptation strength. This design allows larger updates on legally important tokens while preserving more conservative optimization on less informative parts of the output.

\subsection{Policy Optimization Objective}

Given the token-level advantages $A_t$ and token-dependent clipping coefficients $\epsilon_t$, we optimize the policy using a clipped policy gradient objective:
\begin{equation}
\begin{aligned}
\mathcal{L}_{\mathrm{RL}}
&=
\mathbb{E}_t \Big[
\min\big(
r_t A_t,\,
\mathrm{clip}(r_t,1-\epsilon_t,1+\epsilon_t)A_t
\big) \\
&\qquad
-\beta \mathcal{D}_{KL}\big[\pi_{\theta}(a_t \mid s_t)\|\pi_{\theta_{\mathrm{ref}}}(a_t \mid s_t)\big]
\Big].
\end{aligned}
\end{equation}
Here, $r_t$ denotes the probability ratio between the current policy and the old policy.

The objective combines token-level reweighted advantages, token-aware clipping, and KL regularization with respect to a reference policy. The KL term prevents the policy from drifting excessively away from the supervised initialization and helps preserve stable legal-domain generation during post-training.
\begin{table*}[t!]
\centering
\setlength{\tabcolsep}{2.35pt}
\setlength{\dashlinedash}{0.6pt}
\setlength{\dashlinegap}{2.2pt}
\setlength{\arrayrulewidth}{0.45pt}

\resizebox{\linewidth}{!}{
\begin{tabular}{llccccccccccccccc}
\toprule
\multirow{3}{*}{\textbf{Method}} & \multirow{3}{*}{\textbf{Set.}}
& \multicolumn{5}{c}{\textbf{JPO-Dataset}} 
& \multicolumn{5}{c}{\textbf{CAIL2018}} 
& \multicolumn{5}{c}{\textbf{LawBench}} \\
\cmidrule(lr){3-7}\cmidrule(lr){8-12}\cmidrule(lr){13-17}
&& \textbf{Art.} & \textbf{Charge} & \textbf{Sent.} & \textbf{4-Step} & \textbf{Full}
& \textbf{Art.} & \textbf{Charge} & \textbf{Sent.} & \textbf{4-Step} & \textbf{Full}
& \textbf{Art.} & \textbf{Charge} & \textbf{Sent.} & \textbf{4-Step} & \textbf{Full} \\
&& (F1) & (F1) & (Score) & (Comp.) & (Chain)
& (F1) & (F1) & (Score) & (Comp.) & (Chain)
& (F1) & (F1) & (Score) & (Comp.) & (Chain) \\
\midrule


\rowcolor{MTitleGray}\multicolumn{17}{c}{\textbf{Open-Source Results}}\\
\midrule
\multirow{3}{*}{Qwen3-4B-Instruct} 
& Pre-trained & 0.521 & 0.468 & 0.174 & 0.315 & 0.208 & 0.485 & 0.442 & 0.158 & 0.291 & 0.187 & 0.463 & 0.419 & 0.142 & 0.264 & 0.175 \\
& SFT         & 0.884 & 0.858 & 0.405 & 0.902 & 0.652 & 0.856 & 0.831 & 0.381 & 0.882 & 0.627 & 0.834 & 0.809 & 0.364 & 0.857 & 0.598 \\
& \bgB{JPO} & \bgB{\textbf{0.931}} & \bgB{\textbf{0.916}} & \bgB{\textbf{0.542}} & \bgB{\textbf{0.966}} & \bgB{\textbf{0.789}} & \bgB{\textbf{0.903}} & \bgB{\textbf{0.888}} & \bgB{\textbf{0.514}} & \bgB{\textbf{0.951}} & \bgB{\textbf{0.758}} & \bgB{\textbf{0.872}} & \bgB{\textbf{0.861}} & \bgB{\textbf{0.485}} & \bgB{\textbf{0.935}} & \bgB{\textbf{0.724}} \\
\arrayrulecolor{MDash}
\cdashline{1-17}[0.6pt/2.2pt]
\arrayrulecolor{black}

\multirow{3}{*}{Qwen2.5-7B-Instruct} 
& Pre-trained & 0.618 & 0.573 & 0.211 & 0.442 & 0.267 & 0.584 & 0.539 & 0.194 & 0.408 & 0.241 & 0.551 & 0.508 & 0.182 & 0.377 & 0.226 \\
& SFT         & 0.893 & 0.871 & 0.422 & 0.918 & 0.681 & 0.865 & 0.842 & 0.399 & 0.895 & 0.655 & 0.841 & 0.824 & 0.377 & 0.875 & 0.628 \\
& \bgB{JPO} & \bgB{\textbf{0.937}} & \bgB{\textbf{0.928}} & \bgB{\textbf{0.551}} & \bgB{\textbf{0.973}} & \bgB{\textbf{0.806}} & \bgB{\textbf{0.908}} & \bgB{\textbf{0.897}} & \bgB{\textbf{0.521}} & \bgB{\textbf{0.958}} & \bgB{\textbf{0.774}} & \bgB{\textbf{0.879}} & \bgB{\textbf{0.868}} & \bgB{\textbf{0.493}} & \bgB{\textbf{0.942}} & \bgB{\textbf{0.741}} \\
\arrayrulecolor{MDash}
\cdashline{1-17}[0.6pt/2.2pt]
\arrayrulecolor{black}

\multirow{3}{*}{Llama-3-8B-Instruct} 
& Pre-trained & 0.594 & 0.552 & 0.203 & 0.418 & 0.252 & 0.561 & 0.524 & 0.187 & 0.389 & 0.229 & 0.537 & 0.495 & 0.175 & 0.362 & 0.214 \\
& SFT         & 0.871 & 0.847 & 0.396 & 0.897 & 0.641 & 0.846 & 0.822 & 0.375 & 0.872 & 0.614 & 0.823 & 0.798 & 0.354 & 0.851 & 0.585 \\
& \bgB{JPO} & \bgB{\textbf{0.926}} & \bgB{\textbf{0.912}} & \bgB{\textbf{0.534}} & \bgB{\textbf{0.962}} & \bgB{\textbf{0.781}} & \bgB{\textbf{0.897}} & \bgB{\textbf{0.884}} & \bgB{\textbf{0.505}} & \bgB{\textbf{0.946}} & \bgB{\textbf{0.749}} & \bgB{\textbf{0.868}} & \bgB{\textbf{0.856}} & \bgB{\textbf{0.477}} & \bgB{\textbf{0.928}} & \bgB{\textbf{0.716}} \\
\midrule

\rowcolor{MTitleGray}\multicolumn{17}{c}{\textbf{Detailed 3B Analysis}}\\
\midrule
\multirow{12}{*}{Qwen2.5-3B-Instruct} 
& Pre-trained & 0.451 & 0.382 & 0.114 & 0.284 & 0.198 & 0.426 & 0.359 & 0.103 & 0.258 & 0.177 & 0.402 & 0.334 & 0.092 & 0.231 & 0.165 \\
& RL-only & 0.782 & 0.744 & 0.295 & 0.671 & 0.482 & 0.755 & 0.718 & 0.273 & 0.644 & 0.457 & 0.732 & 0.695 & 0.258 & 0.615 & 0.434 \\
& SFT & 0.873 & 0.847 & 0.391 & 0.895 & 0.622 & 0.848 & 0.823 & 0.372 & 0.874 & 0.596 & 0.825 & 0.801 & 0.349 & 0.848 & 0.568 \\
& Vanilla PPO & 0.896 & 0.869 & 0.454 & 0.911 & 0.678 & 0.871 & 0.845 & 0.428 & 0.891 & 0.649 & 0.846 & 0.822 & 0.405 & 0.867 & 0.619 \\
& Legal$\Delta$ & 0.904 & 0.881 & 0.479 & 0.924 & 0.703 & 0.878 & 0.856 & 0.451 & 0.905 & 0.674 & 0.854 & 0.834 & 0.427 & 0.881 & 0.642 \\
& Issue Tree Rubrics & 0.911 & 0.896 & 0.488 & 0.932 & 0.721 & 0.885 & 0.871 & 0.459 & 0.914 & 0.693 & 0.862 & 0.848 & 0.435 & 0.889 & 0.658 \\
& \bgB{JPO} & \bgB{\textbf{0.929}} & \bgB{\textbf{0.921}} & \bgB{\textbf{0.536}} & \bgB{\textbf{0.967}} & \bgB{\textbf{0.791}} & \bgB{\textbf{0.904}} & \bgB{\textbf{0.895}} & \bgB{\textbf{0.505}} & \bgB{\textbf{0.952}} & \bgB{\textbf{0.758}} & \bgB{\textbf{0.877}} & \bgB{\textbf{0.868}} & \bgB{\textbf{0.479}} & \bgB{\textbf{0.932}} & \bgB{\textbf{0.725}} \\
& \bgC{\hspace{3pt}w/o $R_{\mathrm{structure}}$} & \bgC{0.916} & \bgC{0.909} & \bgC{0.512} & \bgC{0.861} & \bgC{0.755} & \bgC{0.892} & \bgC{0.883} & \bgC{0.484} & \bgC{0.835} & \bgC{0.723} & \bgC{0.865} & \bgC{0.857} & \bgC{0.458} & \bgC{0.812} & \bgC{0.693} \\
& \bgC{\hspace{3pt}w/o $R_{\mathrm{consistency}}$} & \bgC{0.918} & \bgC{0.908} & \bgC{0.481} & \bgC{0.958} & \bgC{0.682} & \bgC{0.894} & \bgC{0.882} & \bgC{0.453} & \bgC{0.941} & \bgC{0.653} & \bgC{0.868} & \bgC{0.856} & \bgC{0.428} & \bgC{0.921} & \bgC{0.627} \\
& \bgC{\hspace{3pt}w/o token-level advantage} & \bgC{0.922} & \bgC{0.911} & \bgC{0.519} & \bgC{0.961} & \bgC{0.768} & \bgC{0.898} & \bgC{0.886} & \bgC{0.489} & \bgC{0.945} & \bgC{0.735} & \bgC{0.871} & \bgC{0.859} & \bgC{0.463} & \bgC{0.925} & \bgC{0.704} \\
& \bgC{\hspace{3pt}w/o adaptive clipping} & \bgC{0.924} & \bgC{0.913} & \bgC{0.522} & \bgC{0.963} & \bgC{0.773} & \bgC{0.899} & \bgC{0.888} & \bgC{0.492} & \bgC{0.947} & \bgC{0.741} & \bgC{0.873} & \bgC{0.861} & \bgC{0.467} & \bgC{0.927} & \bgC{0.708} \\
& \bgC{\hspace{3pt}w/ uniform token weights} & \bgC{0.919} & \bgC{0.908} & \bgC{0.501} & \bgC{0.956} & \bgC{0.747} & \bgC{0.895} & \bgC{0.883} & \bgC{0.472} & \bgC{0.938} & \bgC{0.715} & \bgC{0.869} & \bgC{0.857} & \bgC{0.448} & \bgC{0.919} & \bgC{0.685} \\
\midrule

\multirow{12}{*}{Llama-3.2-3B-Instruct} 
& Pre-trained & 0.438 & 0.365 & 0.108 & 0.273 & 0.184 & 0.412 & 0.342 & 0.096 & 0.245 & 0.168 & 0.389 & 0.318 & 0.085 & 0.221 & 0.155 \\
& RL-only & 0.761 & 0.722 & 0.284 & 0.658 & 0.458 & 0.736 & 0.698 & 0.264 & 0.631 & 0.435 & 0.713 & 0.676 & 0.248 & 0.602 & 0.413 \\
& SFT & 0.856 & 0.838 & 0.384 & 0.882 & 0.607 & 0.831 & 0.814 & 0.363 & 0.861 & 0.582 & 0.808 & 0.792 & 0.342 & 0.835 & 0.554 \\
& Vanilla PPO & 0.871 & 0.854 & 0.445 & 0.904 & 0.652 & 0.846 & 0.829 & 0.421 & 0.883 & 0.625 & 0.822 & 0.807 & 0.398 & 0.858 & 0.596 \\
& Legal$\Delta$ & 0.882 & 0.865 & 0.459 & 0.917 & 0.675 & 0.857 & 0.841 & 0.434 & 0.896 & 0.647 & 0.834 & 0.819 & 0.411 & 0.871 & 0.618 \\
& Issue Tree Rubrics & 0.889 & 0.877 & 0.468 & 0.925 & 0.693 & 0.864 & 0.852 & 0.443 & 0.905 & 0.665 & 0.841 & 0.829 & 0.421 & 0.879 & 0.635 \\
& \bgB{JPO} & \bgB{\textbf{0.908}} & \bgB{\textbf{0.895}} & \bgB{\textbf{0.502}} & \bgB{\textbf{0.957}} & \bgB{\textbf{0.755}} & \bgB{\textbf{0.883}} & \bgB{\textbf{0.871}} & \bgB{\textbf{0.474}} & \bgB{\textbf{0.938}} & \bgB{\textbf{0.724}} & \bgB{\textbf{0.859}} & \bgB{\textbf{0.848}} & \bgB{\textbf{0.451}} & \bgB{\textbf{0.912}} & \bgB{\textbf{0.691}} \\
& \bgC{\hspace{3pt}w/o $R_{\mathrm{structure}}$} & \bgC{0.893} & \bgC{0.881} & \bgC{0.478} & \bgC{0.844} & \bgC{0.718} & \bgC{0.868} & \bgC{0.857} & \bgC{0.451} & \bgC{0.821} & \bgC{0.687} & \bgC{0.845} & \bgC{0.834} & \bgC{0.428} & \bgC{0.796} & \bgC{0.658} \\
& \bgC{\hspace{3pt}w/o $R_{\mathrm{consistency}}$} & \bgC{0.895} & \bgC{0.879} & \bgC{0.455} & \bgC{0.941} & \bgC{0.634} & \bgC{0.871} & \bgC{0.855} & \bgC{0.429} & \bgC{0.921} & \bgC{0.605} & \bgC{0.847} & \bgC{0.832} & \bgC{0.407} & \bgC{0.896} & \bgC{0.581} \\
& \bgC{\hspace{3pt}w/o token-level advantage} & \bgC{0.899} & \bgC{0.886} & \bgC{0.485} & \bgC{0.943} & \bgC{0.729} & \bgC{0.875} & \bgC{0.862} & \bgC{0.457} & \bgC{0.924} & \bgC{0.698} & \bgC{0.851} & \bgC{0.839} & \bgC{0.435} & \bgC{0.898} & \bgC{0.669} \\
& \bgC{\hspace{3pt}w/o adaptive clipping} & \bgC{0.901} & \bgC{0.888} & \bgC{0.491} & \bgC{0.945} & \bgC{0.734} & \bgC{0.877} & \bgC{0.864} & \bgC{0.463} & \bgC{0.926} & \bgC{0.704} & \bgC{0.853} & \bgC{0.841} & \bgC{0.441} & \bgC{0.901} & \bgC{0.674} \\
& \bgC{\hspace{3pt}w/ uniform token weights} & \bgC{0.896} & \bgC{0.881} & \bgC{0.478} & \bgC{0.938} & \bgC{0.719} & \bgC{0.872} & \bgC{0.858} & \bgC{0.451} & \bgC{0.919} & \bgC{0.689} & \bgC{0.848} & \bgC{0.835} & \bgC{0.429} & \bgC{0.895} & \bgC{0.661} \\
\bottomrule
\end{tabular}}
\vspace{-4pt}
\caption{Main results on JPO-Dataset, CAIL2018, and LawBench. We report article F1, charge F1, sentence score, 4-Step Completeness, and Full-Chain Consistency.}
\vspace{-12pt}
\label{tab:unified_all}
\end{table*}

\begin{table}[t]
\centering
\scriptsize
\setlength{\tabcolsep}{4pt}
\begin{tabular}{lccc}
\toprule
\rowcolor{MTitleGray}
Dataset & SFT Train & RL Train & Test \\
\midrule
\bgB{JPO-Dataset} & \bgB{239,515} & \bgB{9,691} & \bgB{20,396} \\
CAIL2018 & -- & -- & 30,000 \\
LawBench & -- & -- & 1,500 \\
\bottomrule
\end{tabular}
\vspace{-10pt}
\caption{Dataset statistics used in our experiments.}
\vspace{-16pt}
\label{tab:dataset_stats}
\end{table}

\section{Experiments}
\label{sec:experiments}

\subsection{Experimental Setup}

\paragraph{Datasets.}
We evaluate JPO on three Chinese criminal judgment prediction benchmarks: JPO-Dataset, CAIL2018~\cite{xiao2018cail2018}, and LawBench~\cite{fei2024lawbench}. JPO-Dataset is a newly constructed dataset derived from public judicial documents collected from China Judgments Online, covering the time range of 2024-2026. We introduce it because widely used Chinese criminal judgment prediction benchmarks, e.g., CAIL2018, were built from relatively earlier judicial documents and may not fully reflect more recent criminal case distributions and legal expressions. Following the general construction protocol of CAIL2018, we re-collect recent criminal cases and extract fact descriptions, applicable statutory articles, charges, and sentencing outcomes, together with normalization of article references, charge labels, and sentencing expressions. During construction, data collection, cleaning, and field extraction were carried out jointly with collaborators possessing legal backgrounds. This is essential because raw Chinese judgment documents intermix procedural, factual, and adjudicative components in free text, making them unusable without legal training.
 JPO-Dataset is used for structured supervised fine-tuning and reinforcement learning, while CAIL2018 and LawBench are used as external evaluation benchmarks. Basic statistics are shown in Table~\ref{tab:dataset_stats}.

\paragraph{Models.}
We conduct experiments on five open-source backbones: Qwen2.5-3B/7B-Instruct, Qwen3-4B-Instruct, Llama-3.2-3B-Instruct, and Llama-3-8B-Instruct~\citep{qwen2025qwen25technicalreport, yang2025qwen3technicalreport, grattafiori2024llama3herdmodels}. For the 4B/7B/8B models, we report Pre-trained, SFT, and JPO. For the two representative 3B backbones, we further include RL-only, Vanilla PPO~\citep{schulman2017proximal}, Legal$\Delta$~\citep{dai2025legal}, and Issue Tree Rubrics~\citep{cai2025unilaw}, together with component ablations. The teacher model used for generating structured rationales in SFT is Qwen2.5-72B-Instruct. To verify that JPO's improvements are not tied to this specific teacher, we also conducted a sensitivity analysis using DeepSeek-V2 as an alternative teacher in Appendix~\ref{appendix:different_teacher_model}. We also report several proprietary models as zero-shot reference systems, including DeepSeek-V3.2, Qwen3-32B, GPT-5.2, and Claude-Sonnet-4.5. These systems are included only as reference points for model capability and are not directly comparable to our trained open-source systems because they are evaluated in a zero-shot setting without task-specific post-training on JPO-Dataset.

\paragraph{Evaluation Metrics.}
We report five metrics in Table~\ref{tab:unified_all}. Article prediction and charge prediction are evaluated using macro-F1. Sentence prediction is evaluated by a relative-error-based score:
\begin{equation}
\text{Score}_{\text{sentence}} =
\exp\left(
-\xi \cdot \frac{|\hat{S} - S|}{S}
\right),
\end{equation}
where $\xi = 3$ during evaluation, $\hat{S}$ denotes the predicted sentence, and $S$ denotes the gold sentence. We additionally report \emph{4-Step Completeness}, which measures whether the generated response explicitly contains all four reasoning stages required by our framework, and \emph{Full-Chain Consistency}, which measures whether the generated reasoning remains coherent across fact extraction, statutory analysis, charge determination, and sentence prediction. The latter is computed by averaging the three local consistency scores introduced in Section~3.6. Formal definitions of the two reasoning-oriented metrics are provided in Appendix~\ref{sec:appendix_reasoning_metrics}.

\paragraph{Training Details.}
For structured SFT, we train for 2 epochs with batch size 256. The learning rate is $2\times10^{-5}$ for 3B/4B models and $1\times10^{-5}$ for 7B/8B models. For reinforcement learning, we use learning rate $1\times10^{-6}$, 4 epochs, train batch size 1024, PPO mini-batch size 256, KL loss weight $10^{-3}$, and group size 5. Unless otherwise stated, the reward coefficients are set to $\alpha=0.75$, $\beta=0.0625$, and $\gamma=0.1875$, the token-level advantage scaling factor is $\psi=0.6$, the entropy--logic mixing coefficient is $\zeta=0.5$, and the base clipping coefficient is $\epsilon=0.2$. The choice of $\beta = 0.0625$ is derived from a joint ablation study over the reward weight coefficients reported in Appendix~\ref{sec:appendix_reward_weight_sensitivity}, Table~\ref{tab:reward_sensitivity}, where this combination yields the best overall trade-off among legal prediction, structure completeness, and cross-step consistency. For all JPO variants, reinforcement learning starts from the corresponding SFT checkpoint.

\subsection{Main Results}

Table~\ref{tab:unified_all} reports the unified results. Overall, JPO consistently improves over both the pre-trained and structured SFT baselines across all open-source backbones and all three datasets. The gains appear not only on article prediction, charge prediction, and sentence prediction, but also on 4-Step Completeness and Full-Chain Consistency, indicating that JPO improves both final judgment quality and intermediate reasoning coherence.

Structured SFT already yields large improvements over the pre-trained models, showing that explicit four-step supervision is useful for criminal judgment prediction. JPO still provides clear gains over SFT across all backbones, suggesting that second-stage policy optimization further improves legal grounding and cross-step consistency. The gains are especially clear on compact and mid-sized open-source models, and are particularly consistent on sentence prediction and Full-Chain Consistency.

\begin{figure}[t]
    \centering
    \includegraphics[width=1\linewidth]{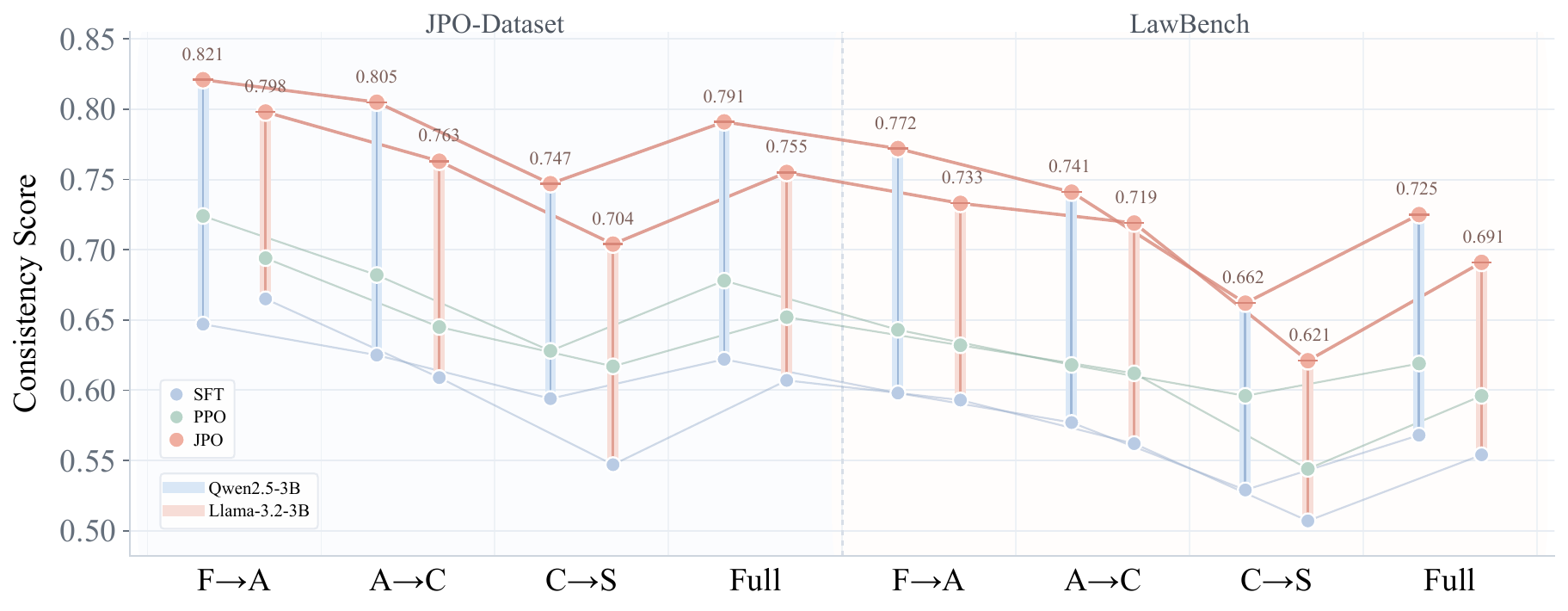}
    \vspace{-12pt}
    \caption{Reasoning-chain breakdown on two representative 3B backbones. F$\rightarrow$A, A$\rightarrow$C, and C$\rightarrow$S denote fact-to-article, article-to-charge, and charge-to-sentence consistency, respectively.}
    \vspace{-14pt}
    \label{tab:reasoning_breakdown}
\end{figure}

\subsection{Detailed Analysis on 3B Backbones}

The lower part of Table~\ref{tab:unified_all} provides a more detailed comparison on two representative 3B backbones. First, the two-stage design is important: RL-only improves over the pre-trained model, but remains clearly weaker than SFT and JPO, indicating that reinforcement learning without structured initialization is insufficient for stable legal reasoning generation. Second, JPO consistently outperforms strong post-training baselines, including Vanilla PPO, Legal$\Delta$, and Issue Tree Rubrics, showing that the gains do not come from reinforcement learning alone, but from the legal-task-specific reward design and token-aware optimization strategy.

Ablation results further show that each component contributes to the final performance. Removing $R_{\mathrm{structure}}$ most strongly affects 4-Step Completeness, while removing $R_{\mathrm{consistency}}$ causes the largest degradation in sentence score and Full-Chain Consistency. Removing token-level advantage reweighting or adaptive clipping also leads to consistent declines, and replacing importance-aware token weights with uniform token weights produces a larger overall drop. These results support both the reward design and the token-aware optimization strategy.

\begin{figure}[t]
    \centering
    \includegraphics[width=1\linewidth]{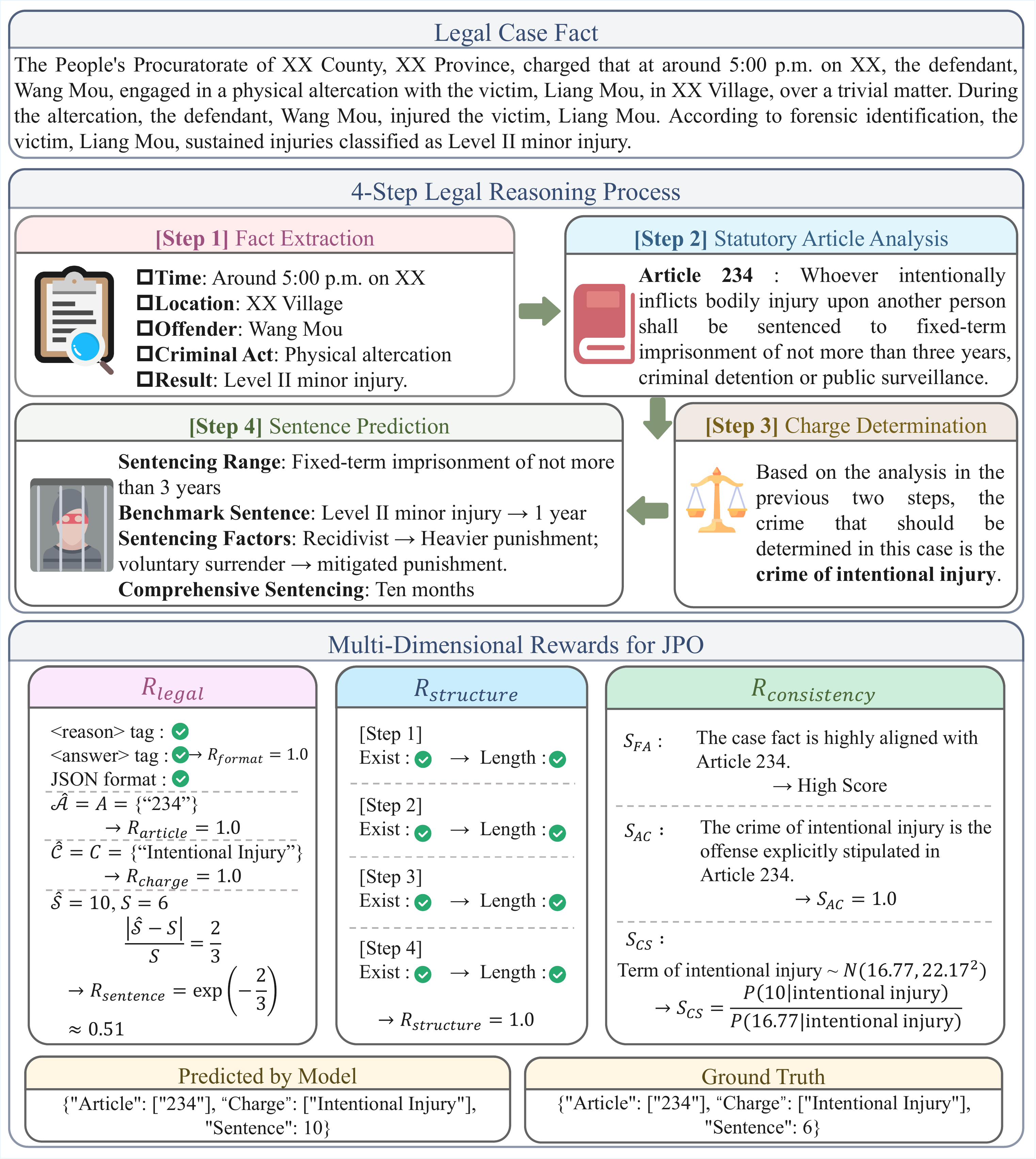}
    \vspace{-12pt}
    \caption{Representative qualitative example illustrating the two-stage process of JPO on a criminal case.}
    \vspace{-14pt}
    \label{fig:case_vis}
\end{figure}

\subsection{Further Analysis}

To better localize the gains, Figure~\ref{tab:reasoning_breakdown} decomposes reasoning quality into three local transitions: fact-to-article consistency, article-to-charge consistency, and charge-to-sentence consistency. Across both 3B backbones, JPO improves all three transitions over SFT and Vanilla PPO, with the largest gains appearing in charge-to-sentence consistency. This pattern is consistent with the stronger improvements on sentence prediction and Full-Chain Consistency in Table~\ref{tab:unified_all}.

Figure~\ref{fig:case_vis} presents a representative qualitative example illustrating the two-stage process of JPO on a criminal case. The figure shows how structured supervised fine-tuning first establishes a four-step reasoning scaffold, and how reinforcement learning further refines the reasoning chain through rewards for legal accuracy, reasoning completeness, and cross-step consistency. Additional qualitative cases and training-dynamics analysis are provided in the appendix.

Appendix~\ref{sec:appendix_extended} reports a set of analyses that probe the results from angles the main table cannot address. These cover the validity of the automatic proxies against expert ratings, a comparison against alternative reasoning decompositions, generalization to multi-charge and multi-defendant cases, format-independent LLM-as-judge and within-subjects expert evaluations, a counterfactual test of reasoning faithfulness, an additive study isolating the contribution of the legal-specific components, and in-context-learning and cost comparisons against proprietary systems.


\section{Conclusion}

We presented \textbf{Juris Policy Optimization (JPO)}, combining four-step supervision, legally structured reinforcement-learning rewards, token-level advantage reweighting, and adaptive clipping. Across multiple open-source backbones and three Chinese legal benchmarks, JPO improves prediction and reasoning metrics over strong baselines, showing the value of optimizing both legal outcomes and the fact--statute--charge--sentence dependency chain.


\section*{Limitations}

JPO is evaluated mainly on single-defendant Chinese criminal cases. It remains stronger than SFT on held-out multi-charge and multi-defendant subsets but degrades when per-defendant attribution is required (Appendix~\ref{sec:appendix_generalization}). Multi-party, civil, procedural, and cross-jurisdictional reasoning remain untested.

Training relies on proxy rewards and gold-aware teacher rationales, while some process metrics mirror JPO's four-step format. We therefore prioritize format-independent outcomes. Expert and template-free judge studies support the findings but do not replace external audits, and teacher rationales may be post-hoc.

Counterfactual tests show improved sensitivity to decisive facts but residual failures remain (Appendix~\ref{sec:appendix_faithfulness}). JPO is not an autonomous legal decision system; deployment requires expert oversight.

\section*{Ethics Statement}

This work studies structured legal reasoning for Chinese criminal judgment prediction using public judicial documents. Although the data are publicly available, they may still contain sensitive information and historical biases.

The structured rationales and reasoning-oriented rewards used in JPO are intended as weak supervision and computational proxies, rather than authoritative legal explanations or substitutes for expert legal judgment.

JPO is developed for legal NLP research and should not be used as an autonomous decision system in real legal practice. Any practical use would require careful human oversight and broader validation.

\bibliography{custom}

\appendix

\newcommand{\apptabstyle}{
  \scriptsize
  \setlength{\tabcolsep}{3pt}
}

\section{Additional Experimental Details}
\label{sec:appendix_additional_results}

This appendix provides additional details for the experimental section, including JPO-Dataset construction, data preprocessing, structured rationale generation, output parsing and evaluation protocol, formal definitions of reasoning-oriented metrics, implementation details of consistency scoring, baseline adaptation, training dynamics, extended ablations, hyper-parameter sensitivity, supplementary results, efficiency analysis, supplementary qualitative examples, and error analysis.

\section{JPO-Dataset Construction}
\label{sec:appendix_dataset}

JPO-Dataset is constructed from criminal judgment documents collected from China Judgments Online. We build this dataset to better reflect more recent criminal case distributions and legal expressions than widely used earlier Chinese criminal judgment prediction benchmarks, which were derived from relatively older public judicial documents.

Following the general construction protocol of CAIL2018, we extract four core fields from each judgment document: fact descriptions, applicable statutory articles, charges, and sentencing outcomes. We then apply preprocessing and normalization to obtain a unified prediction schema across cases.

During preprocessing, we checked the collected public judicial documents for personally identifying information and sensitive details. Consistent with the anonymized format of public Chinese judgment documents, personal names, dates, and locations were either already anonymized or further normalized/redacted into placeholder forms such as “Mou”, “XX”, or “XXX”. We did not intentionally collect offensive content beyond case facts necessary for legal judgment prediction.

\paragraph{Document collection and filtering.}
We collect public criminal cases and retain judgments that contain sufficiently complete factual descriptions and explicit judicial outcomes. We exclude documents with missing key fields, duplicated records, or severely truncated facts. To keep the setting focused and comparable to prior criminal judgment prediction benchmarks, we restrict the current study to single-defendant cases with a normalized fact-to-judgment structure.

\paragraph{Field extraction and normalization.}
For each case, we extract the fact section as model input and use the judicial outcome section to derive the target labels. We normalize article references to unified statutory identifiers, map charge expressions to canonical charge labels, and convert sentencing expressions into normalized sentence values. This step reduces surface-form variation across documents and makes the prediction targets consistent across cases.

\paragraph{Sentence normalization.}
Sentencing outcomes are converted into a normalized scalar value used for both training and evaluation. This normalization mainly resolves surface-form variation in judicial documents, such as different expressions of imprisonment length and equivalent textual formulations. The resulting value is used as the target $S$ in the main paper.

\paragraph{Split usage.}
The resulting dataset is used for both structured supervised fine-tuning and reinforcement learning. The SFT split is used to learn the four-step reasoning template together with the final prediction format, while the RL split is used for post-training under the composite reward. Table~\ref{tab:dataset_stats} in the main paper reports the final split sizes.

\paragraph{Cross-dataset deduplication and leakage control.}
JPO-Dataset and CAIL2018 are both derived from public Chinese judgment documents, so overlap between our training split and the external evaluation sets deserves an explicit check. JPO-Dataset is built from 2024--2026 judgments whereas CAIL2018 derives from substantially earlier documents, and this temporal gap already makes verbatim overlap unlikely. Beyond within-dataset deduplication, we ran a cross-dataset near-duplicate check between the JPO-Dataset training split and both external test sets, using document-level MinHash-LSH over character 5-grams with a Jaccard threshold of 0.8. The check identified 41 near-duplicates against CAIL2018, fewer than 0.02\% of the training data, and none against LawBench. All were removed from training.

\paragraph{Input-length handling.}
For both training and evaluation we cap the input at 2{,}048 tokens and apply tail truncation that preserves the leading content. The same preprocessing and truncation policy is applied to all three datasets. JPO-Dataset facts are short, with a mean of 217.1 and a median of 194 tokens as reported in Table~\ref{tab:jpo_dataset_profile}, so truncation affects fewer than 0.1\% of its cases. Case descriptions in the external benchmarks can be considerably longer: under the same policy, approximately 4.2\% of CAIL2018 and 1.3\% of LawBench cases are truncated.

\paragraph{Artifact release.}
To support reproducibility we release the JPO-Dataset splits, the full training and evaluation code, the teacher prompt shown in Figure~\ref{fig:sft_prompt_template}, the consistency-scoring artifacts (the Naive Bayes article predictor, the article--charge association matrix, and the charge-conditioned sentence statistics), and the additional evaluation subsets and human-evaluation protocol described in Appendix~\ref{sec:appendix_extended}.

\paragraph{Scope and limitations.}
JPO-Dataset is designed for criminal judgment prediction under a structured reasoning formulation. It does not cover all judicial scenarios, such as multi-defendant cases, highly atypical fact patterns, or complex procedural decisions. The dataset therefore serves as a focused benchmark for structured criminal judgment prediction rather than a complete representation of judicial decision-making in practice.

\section{Dataset Profile}
\label{sec:appendix_dataset_profile}

To provide a clearer picture of the evaluation setting, we summarize several basic statistics of JPO-Dataset beyond the split sizes reported in the main paper. These statistics are intended to describe the scale and diversity of the dataset, including the temporal scope of the collected cases, the coverage of charge and article labels, the length of fact descriptions, and the distribution of normalized sentencing outcomes.

JPO-Dataset is constructed from public criminal judgment documents collected over a recent time range. After filtering and normalization, the dataset covers a diverse set of criminal charges and statutory articles, while remaining focused on single-defendant criminal cases. Compared with earlier benchmark-style datasets, its main purpose is not to maximize label coverage, but to provide a cleaner and more recent testbed for structured criminal judgment prediction.

\begin{table}[t]
\centering
\apptabstyle
\begin{tabular}{lc}
\toprule
\rowcolor{MTitleGray}
Statistic & Value \\
\midrule
Time range of collected cases & 2024--2026 \\
Number of unique charges & 192 \\
Number of unique statutory articles & 176 \\
Average fact length (tokens) & 217.1 \\
Median fact length (tokens) & 194 \\
Average number of articles per case & 1.04 \\
Average number of charges per case & 1.02 \\
Average normalized sentence value & 14.0 \\
Median normalized sentence value & 7 \\
\bottomrule
\end{tabular}
\vspace{-6pt}
\caption{Profile statistics of JPO-Dataset.}
\vspace{-10pt}
\label{tab:jpo_dataset_profile}
\end{table}

These statistics help contextualize the task difficulty. In particular, the number of distinct charge and article labels reflects the breadth of the prediction space, while fact length indicates the complexity of the input reasoning context. Sentence statistics provide a coarse description of the target distribution used for the normalized sentencing prediction setting.
\section{Structured Rationale Generation for SFT}
\label{sec:appendix_rationale_generation}

Most legal judgment prediction datasets do not provide explicit intermediate reasoning supervision. To construct structured SFT targets, we use a stronger instruction-following teacher model to generate four-step legal rationales aligned with our reasoning template.

\paragraph{Rationale structure.}
For each case, the teacher is instructed to generate four ordered segments:
(1) fact extraction,
(2) statutory analysis,
(3) charge determination, and
(4) sentence prediction.
Each segment is required to be concise, legally relevant, and aligned with the final normalized answer block.

\paragraph{Target construction.}
The student target consists of the teacher-generated four-step reasoning trace together with the gold judicial outcome. This design allows the model to learn not only the output format, but also a stable intermediate reasoning scaffold before reinforcement learning.

\paragraph{Role of teacher rationales.}
The teacher rationales are used as weak structured supervision rather than as gold legal explanations. Their purpose is to expose the student model to a consistent and interpretable reasoning template. The second-stage reinforcement learning objective then further refines whether the generated reasoning remains correct, complete, and cross-step coherent.

\section{Output Format and Parsing Protocol}
\label{sec:appendix_output_parsing}

Because JPO jointly evaluates final outcomes and intermediate reasoning, we use a unified output schema during both training and evaluation.

\paragraph{Structured response format.}
Each response contains two parts: a four-step reasoning trace and a final prediction block. The reasoning trace is organized into the four predefined stages, while the final prediction block contains normalized article, charge, and sentence predictions.

\paragraph{Parsing of final answers.}
Predicted statutory articles and charges are extracted from the final prediction block and mapped to the normalized label space used by the dataset. Sentence predictions are parsed into the normalized scalar value used for evaluation. Outputs that fail to match the required schema receive a lower format reward and may also produce zero scores for the corresponding set-based or consistency-based terms when parsing fails.

\paragraph{Empty-set handling.}
For reward computation, when the predicted article set $\hat{A}$ is empty, we define the corresponding set-based quantities to be zero. Similarly, when the predicted charge set $\hat{C}$ is empty, the corresponding charge-related reward and consistency terms are set to zero. This avoids undefined behavior in set-based rewards and consistency scores.

\section{Implementation Details of Consistency Scoring}
\label{sec:appendix_consistency_impl}

The main paper defines three local consistency scores: fact-to-article consistency $S_{FA}$, article-to-charge consistency $S_{AC}$, and charge-to-sentence consistency $S_{CS}$. Here we provide additional implementation details.

\paragraph{Fact-to-article consistency.}
The Naive Bayes article predictor used in $S_{FA}$ is trained on the training corpus using fact-side lexical features and article labels. Its role is not to serve as a standalone predictor, but to provide a lightweight estimate of whether the predicted articles are supported by the observed facts.

\paragraph{Article-to-charge consistency.}
The article--charge association matrix used in $S_{AC}$ is built from rule-based associations derived from statutory references and charge annotations in the training data. It is intended as a lightweight compatibility table rather than a complete codification of legal doctrine.

\paragraph{Charge-to-sentence consistency.}
For $S_{CS}$, we estimate a charge-conditioned sentence distribution from the training data. The purpose of this term is to measure whether the predicted sentence lies in a plausible region conditioned on the predicted charge. It should therefore be interpreted as a statistical plausibility signal rather than a normative sentencing model.

\paragraph{Interpretive boundary.}
All three consistency scores are designed as computable proxies that support policy optimization and reasoning-oriented evaluation. They are not intended to fully represent expert legal analysis or replace professional legal judgment.

\section{Baseline Adaptation}
\label{sec:appendix_baseline_adaptation}

To improve experimental fairness, all trainable open-source baselines are adapted to the same underlying prediction setting.

\paragraph{Shared task interface.}
All trainable methods are evaluated under the same structured output schema and the same normalized target space for articles, charges, and sentences.

\paragraph{Initialization and data usage.}
For methods involving post-training, reinforcement learning starts from the corresponding SFT checkpoint unless otherwise noted. This keeps the comparison focused on the effect of the post-training objective rather than differences in initialization.

\paragraph{Interpretation of proprietary references.}
The proprietary systems reported in the main paper are included only as zero-shot references. Since they are not post-trained on JPO-Dataset, they should not be interpreted as directly comparable training baselines.

\section{Definitions of Reasoning-Oriented Metrics}
\label{sec:appendix_reasoning_metrics}

We evaluate reasoning quality with two process-oriented metrics beyond final label accuracy.

\paragraph{4-Step Completeness.}
A generated response is counted as complete if it explicitly contains all four reasoning stages required by our framework: fact extraction, statutory analysis, charge determination, and sentence prediction. In practice, we detect these stages using the structured output template and require each segment to be non-empty. The final 4-Step score is the average completeness rate over the evaluation set.

\paragraph{Full-Chain Consistency.}
Full-Chain Consistency measures the overall coherence of the generated legal reasoning chain. For each response, we first compute three local consistency scores introduced in Section~3.6: fact-to-article consistency $S_{FA}$, article-to-charge consistency $S_{AC}$, and charge-to-sentence consistency $S_{CS}$. We then average these three scores:
\begin{equation}
\mathrm{FullChain}
=
\frac{
S_{FA}+S_{AC}+S_{CS}
}{3}.
\end{equation}
The final Full-Chain score is obtained by averaging this value over all evaluation samples.

\paragraph{Interpretation.}
Article F1, charge F1, and sentence score evaluate the final judgment outcome. By contrast, 4-Step Completeness measures whether the model follows the expected reasoning structure, while Full-Chain Consistency measures whether the reasoning process remains coherent across stages according to the three local consistency proxies defined in the main paper. These metrics are intended to capture useful process signals for structured legal reasoning, rather than to serve as complete substitutes for expert legal evaluation.

\paragraph{Coupling to the training objective.}
Both process metrics are deliberately coupled to JPO's output structure, and Full-Chain Consistency reuses the same proxy signals as the consistency reward, so improvements on them could in principle reflect closer adherence to our own template rather than better legal reasoning. We therefore treat the outcome metrics, which are independent of our format, as the primary evidence, and the process metrics as auxiliary. Three further checks bear on this concern: the robustness of the sentence score to the reward's functional form (Table~\ref{tab:xi_sweep}), the substantial correlation of the consistency proxies with blind expert step-level ratings (Appendix~\ref{sec:appendix_proxy_expert}), and a format-agnostic LLM-as-judge evaluation whose rubric never mentions the four-step template (Appendix~\ref{sec:appendix_llm_judge}). 4-Step Completeness is an unambiguous structural check, so sharing its simple form between reward and metric reflects clarity rather than circular validation.

\section{Training Dynamics}
\label{sec:appendix_training_dynamics}

\begin{figure}[t]
    \centering
    \includegraphics[width=1\linewidth]{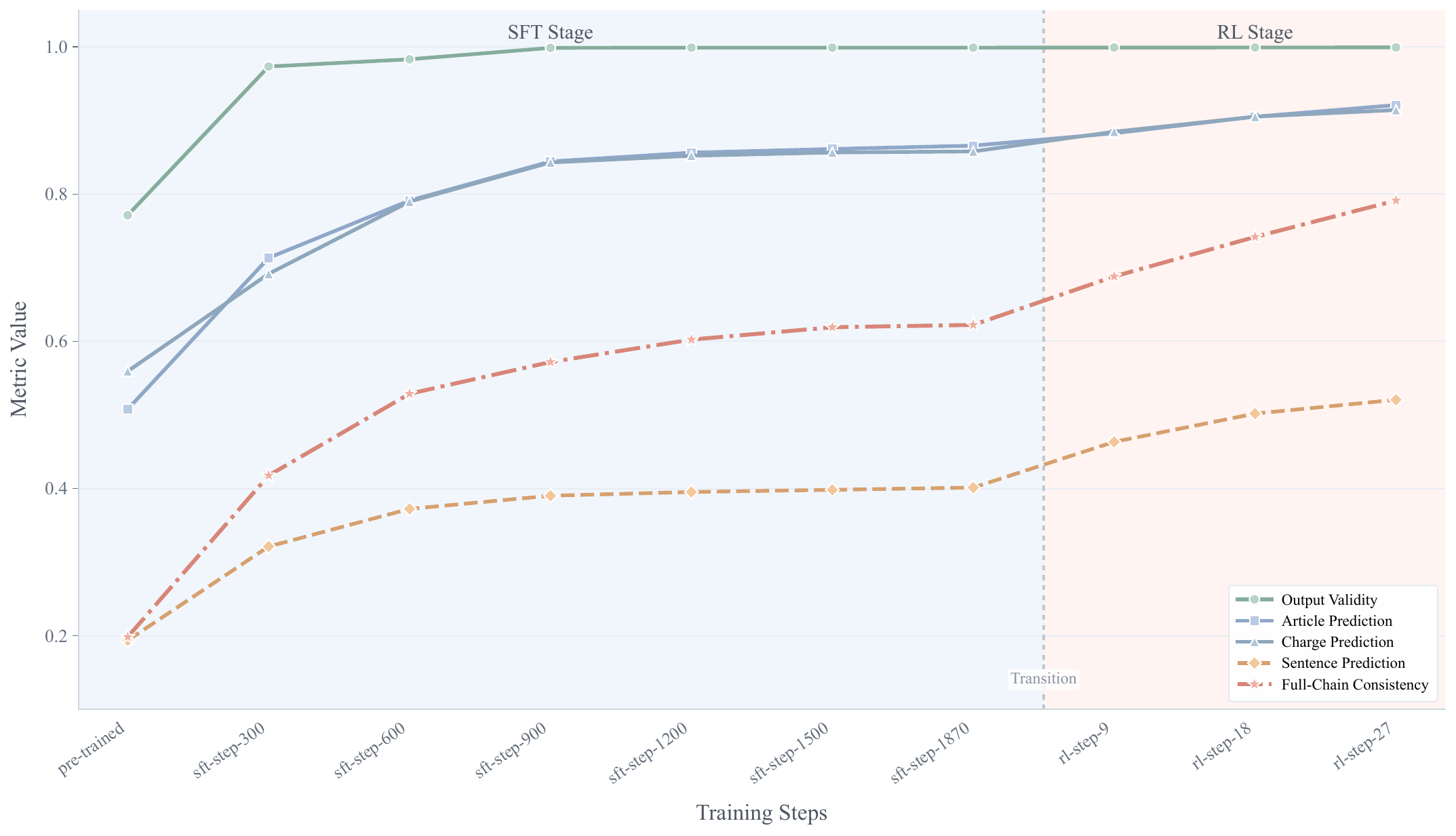}
    \caption{Training dynamics of JPO on the representative Qwen2.5-3B backbone. The SFT stage mainly improves output validity and coarse prediction quality, while the RL stage yields the largest gains in sentence prediction and reasoning consistency.}
    \label{fig:training_curve_appendix}
\end{figure}

Figure~\ref{fig:training_curve_appendix} illustrates how the gains of JPO emerge during training on Qwen2.5-3B. During the SFT stage, the most immediate improvements appear in output validity and coarse legal prediction, reflecting that structured supervision mainly teaches the model to follow the desired reasoning template. During the RL stage, the most visible gains shift to sentence prediction and Full-Chain Consistency, indicating that reinforcement learning primarily improves legal coherence and decision calibration beyond template acquisition alone.

This pattern is consistent with the design of JPO. The first stage mainly stabilizes the reasoning format and teaches the model to expose intermediate legal steps, while the second stage focuses on refining whether these steps remain mutually compatible. In particular, the later gains in sentence prediction suggest that policy optimization is especially useful for improving downstream reasoning quality after the basic structure has already been acquired.

The main paper integrates the core ablation results into the unified main table. Here we further provide a finer-grained ablation analysis for the consistency components on Qwen2.5-3B. The results are shown in Table~\ref{tab:ablation_appendix}. Among the three local consistency transitions, removing charge-to-sentence consistency causes the largest drop in sentence prediction, while removing fact-to-article consistency most directly harms statutory grounding.

These results further support the design intuition of JPO. The three consistency components do not contribute equally: earlier transitions mainly affect whether the model selects legally plausible statutory support, whereas the later charge-to-sentence component plays a more direct role in sentence calibration. The weighting variants also show that combining entropy-based uncertainty with legal-logic-aware importance is more effective than using either signal alone.

\begin{table}[t]
\centering
\apptabstyle
\begin{tabular}{lcccc}
\toprule
\rowcolor{MTitleGray}
Method & Art. F1 & Charge F1 & Sent. & Full-Chain \\
\midrule
\bgB{\textbf{JPO}} & \bgB{\textbf{0.929}} & \bgB{\textbf{0.921}} & \bgB{\textbf{0.536}} & \bgB{\textbf{0.791}} \\
\underline{w/o $S_{FA}$} & \underline{0.923} & \underline{0.915} & \underline{0.520} & \underline{0.768} \\
\underline{w/o $S_{AC}$} & \underline{0.920} & \underline{0.908} & \underline{0.513} & \underline{0.755} \\
\underline{w/o $S_{CS}$} & \underline{0.925} & \underline{0.916} & \underline{0.497} & \underline{0.763} \\
Entropy-only weighting & 0.923 & 0.912 & 0.523 & 0.769 \\
Logic-only weighting & 0.924 & 0.914 & 0.525 & 0.771 \\
w/o KL regularization & 0.925 & 0.915 & 0.515 & 0.741 \\
\bottomrule
\end{tabular}
\vspace{-6pt}
\caption{Extended ablation analysis on Qwen2.5-3B. Bold indicates the full JPO model, and underlined text indicates key ablations.}
\vspace{-10pt}
\label{tab:ablation_appendix}
\end{table}

\section{Extended Hyper-Parameter Sensitivity}
\label{sec:appendix_hyper}

We evaluate the robustness of JPO with respect to several key hyper-parameters on Qwen2.5-3B using JPO-Dataset.

\subsection{Reward Weight Sensitivity}
\label{sec:appendix_reward_weight_sensitivity}
Table~\ref{tab:reward_sensitivity} shows the effect of varying the reward composition. The best overall trade-off is obtained when legal prediction remains the dominant term while structure and consistency provide auxiliary guidance. Over-emphasizing any single component leads to weaker overall performance, suggesting that JPO benefits from balancing final-answer supervision and process-level reasoning supervision.

\begin{table}[t]
\centering
\apptabstyle
\begin{tabular}{cccccc}
\toprule
\rowcolor{MTitleGray}
$\alpha$ & $\beta$ & $\gamma$ & Art. F1 & Charge F1 & Sent. \\
\midrule
0.85 & 0.05   & 0.10   & 0.923 & 0.913 & 0.518 \\
\bgB{\textbf{0.75}} & \bgB{\textbf{0.0625}} & \bgB{\textbf{0.1875}} & \bgB{\textbf{0.929}} & \bgB{\textbf{0.921}} & \bgB{\textbf{0.536}} \\
0.65 & 0.0625 & 0.2875 & 0.926 & 0.917 & 0.530 \\
0.33 & 0.33   & 0.33   & 0.904 & 0.893 & 0.494 \\
\bottomrule
\end{tabular}
\vspace{-6pt}
\caption{Sensitivity to reward weight composition.}
\vspace{-10pt}
\label{tab:reward_sensitivity}
\end{table}

\subsection{Token-Level Advantage Scaling}

Table~\ref{tab:psi_sensitivity} shows the effect of the token-level advantage scaling factor $\psi$. Moderate reweighting performs best, while either removing token-level scaling or applying excessively strong scaling leads to weaker results. This supports the use of importance-aware but still conservative token-level optimization.

\begin{table}[t]
\centering
\apptabstyle
\begin{tabular}{cccc}
\toprule
\rowcolor{MTitleGray}
$\psi$ & Art. F1 & Charge F1 & Sent. \\
\midrule
0.0 & 0.922 & 0.911 & 0.519 \\
0.3 & 0.924 & 0.915 & 0.523 \\
\bgB{\textbf{0.6}} & \bgB{\textbf{0.929}} & \bgB{\textbf{0.921}} & \bgB{\textbf{0.536}} \\
0.9 & 0.927 & 0.918 & 0.531 \\
\bottomrule
\end{tabular}
\vspace{-6pt}
\caption{Sensitivity to token-level advantage scaling $\psi$.}
\vspace{-10pt}
\label{tab:psi_sensitivity}
\end{table}

\subsection{Entropy--Logic Mixing}

Table~\ref{tab:zeta_sensitivity} shows the effect of the entropy--logic mixing coefficient $\zeta$. The best setting lies near the middle of the range, indicating that both uncertainty information and legal-logic-aware stage information contribute to token importance estimation. Using only one of the two signals leads to weaker performance.

\begin{table}[t]
\centering
\apptabstyle
\begin{tabular}{cccc}
\toprule
\rowcolor{MTitleGray}
$\zeta$ & Art. F1 & Charge F1 & Sent. \\
\midrule
0.0  & 0.924 & 0.914 & 0.525 \\
0.25 & 0.926 & 0.917 & 0.529 \\
\bgB{\textbf{0.5}} & \bgB{\textbf{0.929}} & \bgB{\textbf{0.921}} & \bgB{\textbf{0.536}} \\
0.75 & 0.925 & 0.916 & 0.527 \\
1.0  & 0.923 & 0.912 & 0.523 \\
\bottomrule
\end{tabular}
\vspace{-6pt}
\caption{Sensitivity to entropy--logic mixing coefficient $\zeta$.}
\vspace{-10pt}
\label{tab:zeta_sensitivity}
\end{table}

\begin{figure}[t]
\centering
\includegraphics[width=1\linewidth]{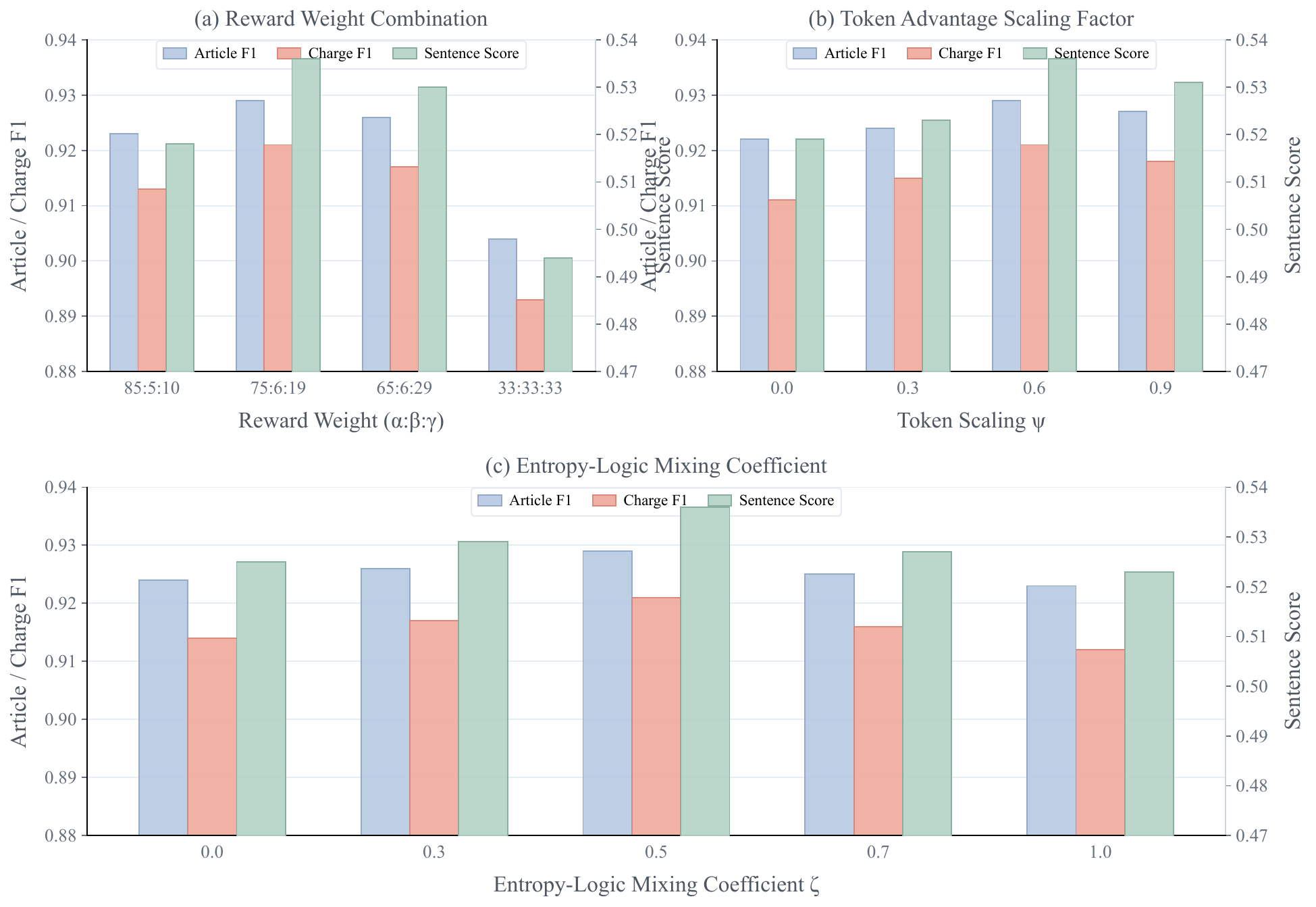}
\caption{Hyper-parameter sensitivity of JPO on the representative Qwen2.5-3B backbone.}
\label{fig:hyper_sensitivity}
\end{figure}

\section{Supplementary Results}

\subsection{Zero-Shot Performance of Proprietary Models}

Table~\ref{tab:proprietary_zero_shot} reports the zero-shot performance of several proprietary models on the three benchmarks. Unlike JPO, these proprietary models were evaluated in a zero-shot setting without any task-specific post-training on JPO-Dataset. Therefore, these results are provided only as reference points for model capability, not as direct comparisons to our trained open-source systems. Appendix~\ref{sec:appendix_icl_cost} extends this comparison with 3-shot in-context learning for both the proprietary models and JPO, together with a per-case cost analysis.

\subsection{Comparison with State-of-the-Art LJP Methods}

To provide a more direct benchmark against prior work, we compare JPO with representative LJP-specific methods from several paradigms: dependency-aware models \citep{zhong-etal-2018-legal, Yang_2019, feng-etal-2022-legal}; structure- and knowledge-enhanced models \citep{xu-etal-2020-distinguish, yue2021neurjudge, liu-etal-2022-augmenting}; retrieval-augmented LLM methods \citep{wu-etal-2023-precedent}; and a recent multi-agent LJP system \citep{liao2026verdictverifiableevolvingreasoning}.

A note on evaluation metrics is necessary. In the main paper, Macro-F1 is computed by averaging per-case F1 scores. However, most prior LJP methods report Macro-F1 computed by averaging per-label F1 scores. To ensure fair comparison, the results reported for JPO in this section follow the per-label averaging convention used by the baselines, rather than the per-case convention used elsewhere in this paper. Sentencing evaluation also varies considerably across papers: classical methods report top-1 accuracy, VERDICT reports Macro-F1, and our submission reports a continuous sentence score. These differences preclude direct absolute comparison on sentencing.

Table~\ref{tab:sota_compare} presents the comparison results on CAIL2018. In terms of Article F1 and Charge F1, JPO substantially outperforms all baseline methods. On article prediction, JPO (Qwen2.5-7B) achieves 90.8, an improvement of 7.5 percentage points over the current SOTA method VERDICT at 83.3. On charge prediction, JPO reaches 89.7, an improvement of 7.3 percentage points over VERDICT at 82.4. Notably, even with the smaller Qwen2.5-3B model, JPO (90.4/89.5) still exceeds VERDICT's multi-agent pipeline on both Article F1 and Charge F1. This demonstrates that JPO, as a single-model post-training framework, achieves stronger performance on core legal judgment prediction tasks. As for sentencing metrics, direct comparison is not feasible due to the different evaluation protocols used across methods, including accuracy, Macro-F1 and continuous score.

\subsection{Pilot Human Evaluation of Reasoning Quality}

\label{appendix:expert_evaluation}

To complement the automatic metrics and provide a preliminary assessment of JPO's generated legal reasoning quality, we conducted a pilot blind expert evaluation. A reviewer with a legal background was invited to evaluate the structured rationales produced by the baseline SFT model and our full JPO framework.

From the JPO-Dataset test set, we randomly sampled 400 cases. For each case, the reviewer was presented with the case facts and the model-generated four-step reasoning trace Fact $\to$ Article $\to$ Charge $\to$ Sentence, with the system identity (SFT vs. JPO) blinded. The reviewer independently judged each of the four reasoning steps as either satisfactory (1) or unsatisfactory (0), resulting in four binary scores per case: Fact, Article, Charge, and Sentence. The JPO model used for this evaluation was Qwen2.5-7B-Instruct, and the baseline was its SFT counterpart. We explicitly note that this is a pilot study with a single legal-background reviewer (n=400) and is not a substitute for a full multi-annotator external expert study, but it provides a meaningful signal of human-perceived quality improvements.

As shown in Table~\ref{tab:pilot_human_eval}, JPO outperforms the SFT baseline across all four reasoning dimensions. The gains are substantial and consistent, with the largest improvement observed on Sentence prediction (+0.22), which aligns with the tendency of SFT models to produce under-calibrated sentencing rationales. The JPO-generated rationales were rated as more legally coherent, better grounded in the cited statutory articles, and more logically connected from facts to the final sentencing decision. Appendix~\ref{sec:appendix_human_eval_large} reports a larger within-subjects study that extends this pilot to three experts, 800 cases, and three systems.

\subsection{Teacher Model Sensitivity Analysis}

\label{appendix:different_teacher_model}

In JPO, Stage I structured rationales are generated by a teacher model. The default teacher used in all main experiments is \textbf{Qwen2.5-72B-Instruct}, chosen as a balance between instruction-following quality and compute (deployable on 4$\times$ A800). To separate teacher quality from framework contribution and verify that JPO's improvements are not tied to a specific teacher, we re-ran the full pipeline with \textbf{DeepSeek-V2} (a larger MoE model) as an alternative teacher. The student model (Qwen2.5-7B-Instruct) and the RL stage configuration were kept identical to the main experiments. All results are reported on the JPO-Dataset.

Table~\ref{tab:teacher_sensitivity} compares the SFT and JPO performance under both teachers. Several observations can be made. First, the stronger teacher (DeepSeek-V2) yields a slightly higher SFT starting point across all metrics, which is expected. Second, JPO consistently improves over SFT under both teachers, with per-metric gains that are comparable (e.g., on Sentence Score, JPO improves by +0.129 under the default teacher and +0.116 under the alternative teacher). The slightly smaller gain under the stronger teacher reflects a mild ceiling effect. Most importantly, the final JPO performance under the two teachers \textbf{converges within $\pm 0.003$} on Article F1, Charge F1, and Sentence Score, demonstrating that JPO's effectiveness is robust to the choice of teacher model. Appendix~\ref{sec:appendix_teacher_scaling} broadens this analysis to five teachers spanning 7B to proprietary scale, and Appendix~\ref{sec:appendix_faithfulness} examines the effect of the teacher \emph{prompt} rather than the teacher model.

\subsection{Sensitivity Analysis of Sentence Reward Coefficient $\xi$}

As noted in the main paper, both the sentence reward $R_{\text{sentence}}$ (Eq. 10) and the sentence evaluation metric $\text{Score}_{\text{sentence}}$ (Eq. 29) adopt the same exponential relative-error form with $\xi = 3$. A legitimate concern is whether JPO's improvements on sentence prediction merely reflect overfitting to the specific functional form of its own reward.

To rule out this possibility, we re-evaluated the same model outputs on the JPO-Dataset under a range of $\xi$ values $\{1, 2, 3, 4, 5\}$. If JPO's gain were tied to the particular choice of $\xi = 3$, its advantage over baselines would diminish or disappear under different $\xi$ values, especially those that are more forgiving ($\xi = 1$) or much stricter ($\xi = 5$).

Table~\ref{tab:xi_sweep} reports the sentence scores for Pre-trained, SFT, and JPO on two representative backbones (Qwen2.5-3B-Instruct and Qwen2.5-7B-Instruct) across all five $\xi$ settings. The results show that JPO maintains a clear and consistent margin over both Pre-trained and SFT baselines at every $\xi$ value. Notably, this includes $\xi = 1$, where the evaluation is substantially more forgiving than the reward's $\xi = 3$, and $\xi = 5$, where the evaluation is much stricter. The relative improvement of JPO over SFT remains stable across the entire sweep.

We therefore conclude that the sentence prediction gains of JPO are not an artifact of matching the specific exponential form or the particular $\xi = 3$ coefficient in its reward. Instead, JPO genuinely improves the calibration and accuracy of sentence predictions, and this improvement generalizes across different evaluation stringency levels.

\begin{table*}[t!]
\centering
\setlength{\tabcolsep}{2.35pt}
\setlength{\dashlinedash}{0.6pt}
\setlength{\dashlinegap}{2.2pt}
\setlength{\arrayrulewidth}{0.45pt}

\resizebox{\linewidth}{!}{
\begin{tabular}{llccccccccccccccc}
\toprule
\multirow{3}{*}{\textbf{Method}} & \multirow{3}{*}{\textbf{Set.}}
& \multicolumn{5}{c}{\textbf{JPO-Dataset}} 
& \multicolumn{5}{c}{\textbf{CAIL2018}} 
& \multicolumn{5}{c}{\textbf{LawBench}} \\
\cmidrule(lr){3-7}\cmidrule(lr){8-12}\cmidrule(lr){13-17}
&& \textbf{Art.} & \textbf{Charge} & \textbf{Sent.} & \textbf{4-Step} & \textbf{Full}
& \textbf{Art.} & \textbf{Charge} & \textbf{Sent.} & \textbf{4-Step} & \textbf{Full}
& \textbf{Art.} & \textbf{Charge} & \textbf{Sent.} & \textbf{4-Step} & \textbf{Full} \\
&& (F1) & (F1) & (Score) & (Comp.) & (Chain)
& (F1) & (F1) & (Score) & (Comp.) & (Chain)
& (F1) & (F1) & (Score) & (Comp.) & (Chain) \\
\midrule

DeepSeek-V3.2 & Zero-Shot & 0.924 & 0.908 & 0.512 & 0.958 & 0.771 & 0.896 & 0.875 & 0.481 & 0.942 & 0.738 & 0.864 & 0.841 & 0.439 & 0.925 & 0.706 \\
Qwen3-32B & Zero-Shot & 0.917 & 0.911 & 0.504 & 0.961 & 0.759 & 0.887 & 0.882 & 0.473 & 0.945 & 0.722 & 0.858 & 0.835 & 0.428 & 0.931 & 0.694 \\
GPT-5.2 & Zero-Shot & 0.912 & 0.894 & 0.489 & 0.947 & 0.744 & 0.879 & 0.861 & 0.455 & 0.933 & 0.705 & 0.845 & 0.829 & 0.412 & 0.916 & 0.683 \\
Claude-Sonnet-4.5 & Zero-Shot & 0.908 & 0.891 & 0.493 & 0.952 & 0.751 & 0.882 & 0.866 & 0.462 & 0.937 & 0.713 & 0.849 & 0.824 & 0.417 & 0.918 & 0.687 \\
\midrule

\bgB{Qwen2.5-3B-Instruct} & 
\bgB{JPO} & \bgB{\textbf{0.929}} & \bgB{\textbf{0.921}} & \bgB{\textbf{0.536}} & \bgB{\textbf{0.967}} & \bgB{\textbf{0.791}} & \bgB{\textbf{0.904}} & \bgB{\textbf{0.895}} & \bgB{\textbf{0.505}} & \bgB{\textbf{0.952}} & \bgB{\textbf{0.758}} & \bgB{\textbf{0.877}} & \bgB{\textbf{0.868}} & \bgB{\textbf{0.479}} & \bgB{\textbf{0.932}} & \bgB{\textbf{0.725}} \\
\bottomrule

\end{tabular}}
\vspace{-8pt}
\caption{Zero-shot performance of proprietary models. These results are for reference only and are not directly comparable to our open-source models trained with task-specific post-training.}
\vspace{-12pt}
\label{tab:proprietary_zero_shot}
\end{table*}

\begin{table}[h!]
\centering
\setlength{\tabcolsep}{2.35pt}
\setlength{\dashlinedash}{0.6pt}
\setlength{\dashlinegap}{2.2pt}
\setlength{\arrayrulewidth}{0.45pt}

\resizebox{\linewidth}{!}{
\label{tab:sota_compare}
\begin{tabular}{lcccc}
\toprule
\rowcolor{MTitleGray}
Method & Paradigm & Art. F1 & Charge F1 & Sentencing (as reported) \\
\midrule
TopJudge & dependency-aware & 0.737 & 0.800 & 0.357 (Acc) \\
MPBFN & dependency-aware & 0.706 & 0.757 & 0.362 (Acc) \\
LADAN & structure \& knowledge & 0.738 & 0.801 & 0.361 (Acc) \\
NeurJudge & structure \& knowledge & 0.797 & 0.807 & 0.374 (Acc) \\
CTM & structure \& knowledge & 0.768 & 0.780 & 0.374 (Acc) \\
EPM & dependency-aware & 0.781 & 0.814 & 0.367 (Acc) \\
PLJP & retrieval + LLM & 0.749 & 0.763 & 0.382 (Acc) \\
VERDICT & multi-agent, SOTA on CAIL2018 & 0.833 & 0.824 & 0.341 (MF1) \\
\midrule
\bgC{JPO (Qwen2.5-3B)} & \bgC{single-model post-training} & \bgC{\underline{0.904}} & \bgC{\underline{0.895}} & \bgC{\underline{0.505 (Sent. Score)}} \\
\bgB{JPO (Qwen2.5-7B)} & \bgB{single-model post-training} & \bgB{\textbf{0.908}} & \bgB{\textbf{0.897}} & \bgB{\textbf{0.521 (Sent. Score)}} \\
\bottomrule
\end{tabular}}
\vspace{-8pt}
\caption{Comparison with SOTA LJP methods on CAIL2018.}
\label{tab:sota_compare}
\end{table}

\begin{table}[htbp]
\centering
\apptabstyle
\setlength{\tabcolsep}{2.35pt}
\setlength{\dashlinedash}{0.6pt}
\setlength{\dashlinegap}{2.2pt}
\setlength{\arrayrulewidth}{0.45pt}

\begin{tabular}{lcccc}
\toprule
\rowcolor{MTitleGray}
\textbf{System} & \textbf{Fact} & \textbf{Article} & \textbf{Charge} & \textbf{Sentence} \\
\midrule
SFT (Qwen2.5-7B-Instruct) & 0.73 & 0.63 & 0.58 & 0.49 \\
\bgB{JPO (Qwen2.5-7B-Instruct)} & \bgB{\textbf{0.87}} & \bgB{\textbf{0.83}} & \bgB{\textbf{0.79}} & \bgB{\textbf{0.71}} \\
\midrule
Improvement & +0.14 & +0.20 & +0.21 & +0.22 \\
\bottomrule
\end{tabular}
\vspace{-8pt}
\caption{Pilot blind expert evaluation results on 400 randomly sampled cases from the JPO-Dataset test set. }
\label{tab:pilot_human_eval}
\end{table}

\begin{table}[htbp]
\centering

\apptabstyle
\setlength{\tabcolsep}{2.35pt}
\setlength{\dashlinedash}{0.6pt}
\setlength{\dashlinegap}{2.2pt}
\setlength{\arrayrulewidth}{0.45pt}

\begin{tabular}{l l c c c}
\toprule
\rowcolor{MTitleGray}
Teacher & Stage & Art. (F1) & Charge (F1) & Sent. (Score) \\
\midrule
\multirow{2}{*}{Qwen2.5-72B-Instruct} & SFT & 0.893 & 0.871 & 0.422 \\
 & \bgB{JPO} & \bgB{\textbf{0.937}} & \bgB{\textbf{0.928}} & \bgB{\textbf{0.551}} \\
\midrule
\multirow{2}{*}{DeepSeek-V2} & SFT & 0.901 & 0.875 & 0.438 \\
 & \bgB{JPO} & \bgB{\textbf{0.939}} & \bgB{\textbf{0.926}} & \bgB{\textbf{0.554}} \\
\bottomrule
\end{tabular}
\vspace{-8pt}
\caption{Teacher model sensitivity analysis on JPO-Dataset.}
\label{tab:teacher_sensitivity}
\end{table}

\begin{table}[htbp]
\centering

\apptabstyle
\setlength{\tabcolsep}{2.35pt}
\setlength{\dashlinedash}{0.6pt}
\setlength{\dashlinegap}{2.2pt}
\setlength{\arrayrulewidth}{0.45pt}

\begin{tabular}{lccccc}
\toprule
\rowcolor{MTitleGray}
Model & $\xi$ &  Pre-trained & SFT & JPO \\
\midrule
\multirow{5}{*}{Qwen2.5-3B-Instruct} & 1.0 & 0.317 & 0.613 & \textbf{0.747} \\
 & 2.0 & 0.185 & 0.472 & \textbf{0.614} \\
 & 3.0 & 0.114 & 0.391 & \textbf{0.536} \\
 & 4.0 & 0.086 & 0.347 & \textbf{0.475} \\
 & 5.0 & 0.067 & 0.305 & \textbf{0.437} \\
\cmidrule{1-5}
\multirow{5}{*}{Qwen2.5-7B-Instruct} & 1.0 & 0.430 & 0.657 & \textbf{0.765} \\
 & 2.0 & 0.286 & 0.513 & \textbf{0.633} \\
 & 3.0 & 0.211 & 0.422 & \textbf{0.551} \\
 & 4.0 & 0.174 & 0.373 & \textbf{0.491} \\
 & 5.0 & 0.149 & 0.331 & \textbf{0.452} \\
\bottomrule
\end{tabular}
\vspace{-8pt}
\caption{Sentence score sensitivity to $\xi$ on JPO-Dataset.}
\label{tab:xi_sweep}
\end{table}

\section{Training Efficiency and Stability}
\label{sec:appendix_efficiency}

Table~\ref{tab:efficiency} reports the efficiency and stability comparison on Qwen2.5-3B. JPO introduces moderate additional training cost relative to SFT, but yields substantially stronger reasoning quality and lower variation than reward-only PPO.

The efficiency results suggest that the additional complexity of JPO is practically manageable: compared with vanilla PPO, the increase in GPU hours is limited, while the gains in reasoning quality and stability are substantial. This trade-off is favorable for legal-domain post-training, where stable structured generation is often more important than marginal reductions in training cost.

\begin{table}[t]
\centering
\apptabstyle
\begin{tabular}{lccc}
\toprule
\rowcolor{MTitleGray}
Method & GPU Hours & Avg. Length & Std. Full-Chain \\
\midrule
SFT & 34 & 456 & 0.018 \\
Vanilla PPO & 49 & 471 & 0.025 \\
\bgB{\textbf{JPO}} & \bgB{\textbf{53}} & \bgB{\textbf{478}} & \bgB{\textbf{0.015}} \\
\bottomrule
\end{tabular}
\vspace{-6pt}
\caption{Training efficiency and stability on Qwen2.5-3B.}
\vspace{-10pt}
\label{tab:efficiency}
\end{table}

\pagebreak[2]
\begin{table}[!htb]
\centering
\apptabstyle
\begin{tabular}{lc}
\toprule
\rowcolor{MTitleGray}
Error Type & Proportion \\
\midrule
Missing secondary article & 27.5\% \\
Correct charge but wrong sentence band & 25.3\% \\
Confusion between neighboring charges & 21.6\% \\
Incomplete mitigation/aggravation reasoning & 16.7\% \\
Formatting or parsing errors & 8.9\% \\
\bottomrule
\end{tabular}
\vspace{-6pt}
\caption{Main remaining error types of JPO.}
\vspace{-10pt}
\label{tab:error_types}
\end{table}

\section{Additional Qualitative Examples}
\label{sec:appendix_more_cases}

In the main paper, we use a compact flow-style figure to visualize a representative case. Here we provide additional qualitative observations based on detailed generated outputs.

\paragraph{Example A: Better statutory grounding.}
In repeated narcotics-trafficking cases, the SFT model often predicts the correct charge but gives only shallow statutory justification. JPO more reliably connects trafficking frequency, drug quantity, and statutory threshold to the cited articles, yielding a better-aligned fact-to-article transition.

\paragraph{Example B: Improved sentence calibration.}
In theft and fraud cases with confession but limited restitution, SFT tends to under-calibrate the sentence. JPO more often predicts a sentence that falls in a plausible charge-conditioned region and better explains why mitigation is limited, producing a more coherent charge-to-sentence transition.

\paragraph{Example C: Stronger multi-stage coherence.}
In cases with highly similar facts but different legal thresholds, JPO is more likely than SFT to preserve coherence from fact extraction to statute selection and then to charge determination, instead of jumping directly to the final answer.

\paragraph{Interpretive caution.}
These qualitative examples are intended to illustrate the types of improvements encouraged by JPO, rather than to serve as formal legal evaluations. They should therefore be interpreted as complementary evidence alongside the quantitative metrics reported in the main paper.

\section{Error Analysis}
\label{sec:appendix_error}

Table~\ref{tab:error_types} summarizes the main remaining error types of JPO. The most common failures involve secondary article omission, difficult sentence calibration under fine-grained aggravating and mitigating factors, and confusion between neighboring charges with similar fact patterns.

The error distribution suggests that the main remaining challenge is no longer gross formatting failure, but fine-grained legal distinction. In particular, secondary article omission and difficult sentence calibration indicate that even when the model captures the broad reasoning chain correctly, it can still miss more subtle statutory interactions or sentencing adjustments. This observation is consistent with the fact that later-stage legal reasoning remains the hardest part of the task.

\section{Extended Validation and Generalization Analyses}
\label{sec:appendix_extended}

The analyses in this section examine JPO from four complementary angles that the main results do not settle on their own: whether the automatic consistency proxies track expert legal judgment (Appendix~\ref{sec:appendix_proxy_expert}), whether the four-step decomposition and its stage-level token weighting are the right design choices (Appendices~\ref{sec:appendix_structure_compare} and~\ref{sec:appendix_logic_granularity}), whether the reported reasoning gains survive evaluation protocols that are independent of JPO's own reward and output format (Appendices~\ref{sec:appendix_llm_judge}--\ref{sec:appendix_faithfulness}), and how the framework behaves outside its training regime, across teachers, and against proprietary systems (Appendices~\ref{sec:appendix_generalization}, \ref{sec:appendix_icl_cost}, and~\ref{sec:appendix_teacher_scaling}).

\subsection{Correlation Between Automatic Proxies and Expert Ratings}
\label{sec:appendix_proxy_expert}

The consistency rewards of JPO are computable proxies rather than expert legal analysis, which raises the question of whether they track genuine reasoning quality at all. To measure this directly, we reuse the step-level expert ratings collected in the blind within-subjects study described in Appendix~\ref{sec:appendix_human_eval_large}, in which three legal experts rated the reasoning of each system on 800 randomly sampled JPO-Dataset test cases with system identity hidden. We correlate each automatic consistency score against the corresponding expert rating of the reasoning step it is meant to assess.

Table~\ref{tab:proxy_expert_corr} reports the results for Qwen2.5-7B-Instruct. Every proxy shows substantial rank correlation with expert judgment, ranging from $\rho = 0.64$ for charge-to-sentence consistency to $\rho = 0.72$ for the aggregate Full-Chain score. The proxies therefore capture real signal about reasoning quality, while the fact that no correlation approaches unity confirms that they remain approximations rather than a complete account of legal reasoning. The comparatively lower correlation of $S_{CS}$ is expected, since sentencing depends on case-specific mitigating and aggravating circumstances that a charge-conditioned statistical prior cannot fully represent.

\begin{table}[H]
\centering
\apptabstyle
\begin{tabular}{llc}
\toprule
\rowcolor{MTitleGray}
Automatic proxy & Expert step judged & Spearman $\rho$ \\
\midrule
$S_{FA}$ (fact-to-article) & Statutory analysis & 0.67 \\
$S_{AC}$ (article-to-charge) & Charge determination & 0.71 \\
$S_{CS}$ (charge-to-sentence) & Sentence prediction & 0.64 \\
\bgB{Full-Chain Consistency} & \bgB{Overall reasoning} & \bgB{\textbf{0.72}} \\
\bottomrule
\end{tabular}
\vspace{-6pt}
\caption{Correlation between the automatic consistency proxies and blind expert step-level ratings, computed on the 800-case expert study of Appendix~\ref{sec:appendix_human_eval_large} (Qwen2.5-7B-Instruct).}
\vspace{-10pt}
\label{tab:proxy_expert_corr}
\end{table}

\subsection{Comparison of Alternative Reasoning Decompositions}
\label{sec:appendix_structure_compare}

The four-step decomposition follows the adjudication workflow of Chinese criminal law and admits a clean operationalization of cross-step consistency over its three transitions. To test whether this conceptual motivation translates into an empirical advantage, we re-ran the complete JPO pipeline under four reasoning schemes, keeping the SFT and RL configuration identical and adapting only the reward definitions to each structure: free-form chain-of-thought, a three-step fact--element--charge scheme in which sentencing is folded into the charge step, an issue-tree decomposition, and our four-step scheme.

As Table~\ref{tab:structure_compare} shows, all structured schemes outperform free-form CoT, and the four-step scheme is strongest overall. The margin is widest on sentence prediction and Full-Chain Consistency, which is precisely what the design predicts: only the four-step scheme exposes an explicit sentencing-reasoning stage over which consistency can be supervised. We do not claim universal optimality. The gains should be largest for tasks whose adjudication is naturally organized as this dependency chain, and transferring JPO to a different structure mainly requires redefining the corresponding consistency relations and rewards, while the two-stage optimization machinery remains unchanged.

\begin{table}[H]
\centering
\apptabstyle
\begin{tabular}{lcccc}
\toprule
\rowcolor{MTitleGray}
Reasoning structure & Art. F1 & Charge F1 & Sent. & Full-Chain \\
\midrule
Standard CoT (free-form) & 0.921 & 0.907 & 0.503 & 0.724 \\
Fact--Element--Charge (3-step) & 0.928 & 0.914 & 0.517 & 0.749 \\
Issue-tree decomposition & 0.931 & 0.918 & 0.525 & 0.768 \\
\bgB{Four-step F$\to$A$\to$C$\to$S} & \bgB{\textbf{0.937}} & \bgB{\textbf{0.928}} & \bgB{\textbf{0.551}} & \bgB{\textbf{0.806}} \\
\bottomrule
\end{tabular}
\vspace{-6pt}
\caption{Reasoning-structure comparison under the JPO pipeline (Qwen2.5-7B-Instruct, JPO-Dataset). Rewards are adapted to each structure; all other settings are identical.}
\vspace{-10pt}
\label{tab:structure_compare}
\end{table}

\subsection{Generalization to Multi-Charge and Multi-Defendant Cases}
\label{sec:appendix_generalization}

JPO-Dataset is restricted to single-defendant cases, and its profile statistics in Table~\ref{tab:jpo_dataset_profile} show roughly one article and one charge per case. The main results therefore speak mostly to regular, near-single-charge adjudication. To probe generalization beyond this regime, we curated two harder evaluation subsets from newly collected held-out judgments outside JPO-Dataset, again with our legal collaborators: a multi-charge subset of 1{,}000 cases and a multi-defendant subset of 800 cases. We evaluated the same Qwen2.5-7B-Instruct SFT and JPO checkpoints on both subsets without any retraining.

Table~\ref{tab:complex_generalization} reports the outcome. Both models degrade substantially on the harder subsets, most sharply on multi-defendant cases, which lie entirely outside the single-defendant training regime and require per-defendant attribution that the models never observed. Even so, JPO retains a clear margin over SFT on every metric, improving sentence score by 0.104 on the multi-charge subset and by 0.105 on the multi-defendant subset. Structured post-training therefore transfers to harder case types rather than collapsing, while the large absolute drop confirms genuine headroom. Extending JPO to fuller multi-party coverage would require per-defendant reasoning chains and charge-level consistency terms, which we leave to future work.

\begin{table}[H]
\centering
\apptabstyle
\resizebox{\linewidth}{!}{
\begin{tabular}{llcccc}
\toprule
\rowcolor{MTitleGray}
Evaluation subset & Model & Art. F1 & Charge F1 & Sent. & Full-Chain \\
\midrule
\multirow{2}{*}{Single-def./single-charge} & SFT & 0.893 & 0.871 & 0.422 & 0.681 \\
 & \bgB{JPO} & \bgB{\textbf{0.937}} & \bgB{\textbf{0.928}} & \bgB{\textbf{0.551}} & \bgB{\textbf{0.806}} \\
\midrule
\multirow{2}{*}{Multi-charge ($n$=1,000)} & SFT & 0.812 & 0.774 & 0.348 & 0.579 \\
 & \bgB{JPO} & \bgB{\textbf{0.869}} & \bgB{\textbf{0.836}} & \bgB{\textbf{0.452}} & \bgB{\textbf{0.695}} \\
\midrule
\multirow{2}{*}{Multi-defendant ($n$=800)} & SFT & 0.701 & 0.643 & 0.246 & 0.437 \\
 & \bgB{JPO} & \bgB{\textbf{0.783}} & \bgB{\textbf{0.724}} & \bgB{\textbf{0.351}} & \bgB{\textbf{0.558}} \\
\bottomrule
\end{tabular}}
\vspace{-6pt}
\caption{Generalization to complex cases (Qwen2.5-7B-Instruct, no retraining). The two harder subsets are drawn from held-out judgments outside JPO-Dataset.}
\vspace{-10pt}
\label{tab:complex_generalization}
\end{table}

\subsection{Granularity of the Legal-Logic Token Weight}
\label{sec:appendix_logic_granularity}

The legal-logic component of the token weight $L_t$ is assigned at the reasoning-stage level: all tokens in the statutory analysis segment share the fact-to-article consistency signal, all tokens in the charge determination segment share the article-to-charge signal, and so on. This design does not pinpoint which individual tokens are correct or erroneous. To quantify what is lost by not going finer, we compared four granularities on Qwen2.5-3B-Instruct, ranging from an entropy-only baseline with no logic weight to a genuinely token-level variant driven by a legal-entailment model that scores each token's local support.

Table~\ref{tab:logic_granularity} shows that stage-level weighting captures most of the benefit of the legal-logic weight relative to the entropy-only baseline, while finer granularities add at most 0.008 sentence score at substantially higher training cost. The proposed design is therefore best understood as an efficient middle ground rather than a claim of fine-grained token-level correctness supervision. Element-level annotation or entailment-based scoring remains the natural route to genuinely finer legal supervision.

\begin{table}[H]
\centering
\apptabstyle
\resizebox{\linewidth}{!}{
\begin{tabular}{lccccc}
\toprule
\rowcolor{MTitleGray}
Logic-weight granularity & Art. F1 & Charge F1 & Sent. & Full-Chain & Cost \\
\midrule
Entropy-only (no logic weight) & 0.923 & 0.912 & 0.523 & 0.769 & 1.00$\times$ \\
\bgB{Stage-level (proposed)} & \bgB{0.929} & \bgB{0.921} & \bgB{0.536} & \bgB{0.791} & \bgB{1.05$\times$} \\
Span-level (per-sentence) & 0.931 & 0.923 & 0.540 & 0.797 & 1.4$\times$ \\
Token-level (entailment model) & 0.933 & 0.925 & 0.544 & 0.802 & 2.6$\times$ \\
\bottomrule
\end{tabular}}
\vspace{-6pt}
\caption{Granularity of the legal-logic token weight (Qwen2.5-3B-Instruct, JPO-Dataset). Cost is training time relative to the entropy-only baseline.}
\vspace{-10pt}
\label{tab:logic_granularity}
\end{table}

\subsection{Format-Independent LLM-as-Judge Evaluation}
\label{sec:appendix_llm_judge}

4-Step Completeness and Full-Chain Consistency are both coupled to JPO's output structure, so improvements on them could in principle reflect closer adherence to our own template rather than better legal reasoning. The outcome metrics already provide one independent check, since article F1, charge F1, and sentence score are format-agnostic and JPO improves all three on all datasets. As a second check, we added a reasoning-quality signal that is deliberately format-agnostic: GPT-5.2 scores each rationale from 1 to 5 for legal soundness, factual grounding, and logical coherence, given only the case facts and the free-text rationale, under a rubric that never mentions our four-step template.

Table~\ref{tab:llm_judge} reports the result on 500 cases with Qwen2.5-3B-Instruct backbones. JPO ranks highest under a judge that is agnostic to its output structure, and a parallel run with Claude-Sonnet-4.5 as judge produced the same ordering. Together with the proxy--expert correlations in Table~\ref{tab:proxy_expert_corr}, this addresses the coupling concern from an angle independent of both the reward and the format. We treat the outcome metrics as the primary evidence throughout the paper and the process metrics as auxiliary.

\begin{table}[H]
\centering
\apptabstyle
\begin{tabular}{lcc}
\toprule
\rowcolor{MTitleGray}
Model & LLM-judge rating (1--5) & Win rate vs.\ SFT \\
\midrule
SFT & 3.14 & -- \\
Vanilla PPO & 3.36 & 57\% \\
Issue Tree Rubrics & 3.53 & 63\% \\
\bgB{JPO} & \bgB{\textbf{3.90}} & \bgB{\textbf{74\%}} \\
\bottomrule
\end{tabular}
\vspace{-6pt}
\caption{Format-independent LLM-as-judge evaluation (Qwen2.5-3B-Instruct, 500 cases, judge = GPT-5.2). The judging rubric does not mention the four-step template.}
\vspace{-10pt}
\label{tab:llm_judge}
\end{table}

\subsection{Within-Subjects Expert Evaluation}
\label{sec:appendix_human_eval_large}

The pilot study in Appendix~\ref{appendix:expert_evaluation} used a single reviewer on 400 cases. To obtain a stronger human signal that is fully independent of the reward, we conducted a larger within-subjects study. We sampled 800 cases from the JPO-Dataset test set and, for each case, presented the four-step traces of the pretrained, SFT, and full JPO models, all built on Qwen2.5-7B-Instruct. Three legal experts each rated all three systems on all 800 cases, with system identity hidden and the order of the three traces randomized per case, giving a binary score per reasoning step.

Table~\ref{tab:human_eval_within} reports the per-system scores averaged over the three experts. Inter-annotator agreement was Fleiss' $\kappa = 0.73$, which is substantial, so the ranking reflects genuine quality differences rather than scorer leniency. Because the protocol is shared across systems, blinded, order-randomized, and independent of the reward, the consistent superiority of JPO corroborates that its reasoning gains are real. The gap is widest on sentence prediction, matching the pattern observed in the automatic metrics.

\begin{table}[H]
\centering
\apptabstyle
\begin{tabular}{lcccc}
\toprule
\rowcolor{MTitleGray}
Model & Fact & Article & Charge & Sentence \\
\midrule
Pretrained & 0.62 & 0.51 & 0.49 & 0.28 \\
SFT & 0.71 & 0.66 & 0.59 & 0.47 \\
\bgB{JPO} & \bgB{\textbf{0.88}} & \bgB{\textbf{0.81}} & \bgB{\textbf{0.77}} & \bgB{\textbf{0.70}} \\
\bottomrule
\end{tabular}
\vspace{-6pt}
\caption{Within-subjects expert evaluation on Qwen2.5-7B-Instruct, averaged over 3 experts $\times$ 800 cases. Fleiss' $\kappa = 0.73$.}
\vspace{-10pt}
\label{tab:human_eval_within}
\end{table}

\subsection{Faithfulness of Teacher-Generated Rationales}
\label{sec:appendix_faithfulness}

Stage-I rationales are produced by a teacher that observes both the case and the gold judgment, which risks training the student to produce plausible rationalizations of the gold label rather than reasoning that causally drives the prediction. We observed exactly this failure in early experiments, where an unconstrained teacher produced backward, label-first explanations. We report two analyses that bear on faithfulness.

\paragraph{Teacher-prompt ablation.} We regenerated the Stage-I rationales with an unconstrained answer-first prompt and retrained with everything else held fixed. Table~\ref{tab:teacher_prompt_ablation} shows that the forward-constrained prompt used throughout the paper yields both better downstream sentence prediction and markedly higher counterfactual consistency, so the constraint is doing substantive work rather than merely tidying the output.

\paragraph{Counterfactual consistency.} On 500 test cases, our legal collaborators edited a single decision-relevant fact whose change should flip the outcome, such as injury grade, drug quantity, or amount involved. We then measured the counterfactual consistency rate, defined as the fraction of cases in which the model's cited article, charge, or sentence updates in the legally expected direction. A causally faithful reasoner should update; a post-hoc rationalizer often will not. Table~\ref{tab:counterfactual} shows that JPO reaches 0.83 against 0.61 for SFT and 0.43 for the pretrained model. JPO's reasoning is thus substantially more responsive to the facts that legally determine the outcome, which is evidence that the learned traces drive the prediction rather than decorate it.

\begin{table}[H]
\centering
\apptabstyle
\resizebox{\linewidth}{!}{
\begin{tabular}{lccc}
\toprule
\rowcolor{MTitleGray}
Teacher prompt & SFT Sent. & JPO Sent. & JPO counterfact. \\
\midrule
Unconstrained (answer-first) & 0.404 & 0.522 & 0.60 \\
\bgB{Constrained forward (ours)} & \bgB{\textbf{0.422}} & \bgB{\textbf{0.551}} & \bgB{\textbf{0.83}} \\
\bottomrule
\end{tabular}}
\vspace{-6pt}
\caption{Teacher-prompt ablation (Qwen2.5-7B-Instruct, JPO-Dataset). The last column reports the counterfactual consistency rate of the resulting JPO model.}
\vspace{-10pt}
\label{tab:teacher_prompt_ablation}
\end{table}

\begin{table}[H]
\centering
\apptabstyle
\begin{tabular}{lc}
\toprule
\rowcolor{MTitleGray}
Model & Counterfactual consistency rate \\
\midrule
Pretrained & 0.43 \\
SFT & 0.61 \\
\bgB{JPO} & \bgB{\textbf{0.83}} \\
\bottomrule
\end{tabular}
\vspace{-6pt}
\caption{Counterfactual consistency rate (Qwen2.5-7B-Instruct, 500 edited cases, constrained-prompt models).}
\vspace{-10pt}
\label{tab:counterfactual}
\end{table}

\subsection{From a Generic RL Recipe to JPO}
\label{sec:appendix_additive}

The individual ingredients of JPO -- structured CoT supervision, reward shaping, PPO-style optimization, token weighting, and adaptive clipping -- are each familiar. Our claim is that their legal-task-specific adaptation is what delivers the gains, not the generic recipe. To make this concrete, we ran an additive study on Qwen2.5-3B-Instruct that starts from the structured SFT checkpoint and switches on one component at a time, moving from a generic RL setup to the full legal-specific design.

Table~\ref{tab:additive_study} isolates where the improvement originates. The two legal-specific steps drive it: the legal composite reward adds 0.035 sentence score and 0.057 Full-Chain, and legal-logic-aware token optimization adds a further 0.035 and 0.044. Generic entropy-only token weighting, by contrast, contributes little on its own, adding only 0.012 sentence score. The value of JPO therefore lies in the task-specific adaptation rather than in assembling off-the-shelf components.

\begin{table}[H]
\centering
\apptabstyle
\resizebox{\linewidth}{!}{
\begin{tabular}{lcccc}
\toprule
\rowcolor{MTitleGray}
Configuration & Art. F1 & Charge F1 & Sent. & Full-Chain \\
\midrule
Structured four-step SFT & 0.873 & 0.847 & 0.391 & 0.622 \\
+ generic PPO (outcome-only reward) & 0.896 & 0.869 & 0.454 & 0.678 \\
+ generic entropy-only token weighting & 0.900 & 0.875 & 0.466 & 0.690 \\
+ legal composite reward & 0.919 & 0.908 & 0.501 & 0.747 \\
\bgB{JPO (+ legal-logic token opt.)} & \bgB{\textbf{0.929}} & \bgB{\textbf{0.921}} & \bgB{\textbf{0.536}} & \bgB{\textbf{0.791}} \\
\bottomrule
\end{tabular}}
\vspace{-6pt}
\caption{From a generic RL recipe to JPO (Qwen2.5-3B-Instruct, JPO-Dataset). The legal composite reward combines the structure and consistency terms.}
\vspace{-10pt}
\label{tab:additive_study}
\end{table}

\subsection{In-Context Learning and Deployment Cost}
\label{sec:appendix_icl_cost}

Table~\ref{tab:proprietary_zero_shot} compares JPO against proprietary systems under zero-shot evaluation. Two natural follow-up questions are whether demonstrations close the gap and whether the cost of post-training is justified relative to simply querying an API.

Table~\ref{tab:icl} extends the comparison with 3-shot in-context learning for both the proprietary models and the JPO model itself. Demonstrations barely move the proprietary models and slightly hurt some metrics, and they add almost nothing to JPO, which is consistent with JPO having already internalized the structured reasoning that demonstrations are meant to induce. Even with demonstrations, both proprietary models remain below JPO on nearly every metric despite a parameter gap of more than an order of magnitude.

Table~\ref{tab:cost} reports the corresponding cost picture. At deployment scale the JPO model is both more accurate and far cheaper per case than API calls, so the one-time training cost is quickly amortized. These figures are indicative rather than exact: the API entries apply each provider's public list pricing to the average tokens per case, while the local entry amortizes A800 GPU time over inference and excludes the one-time training cost reported separately.

\begin{table}[H]
\centering
\apptabstyle
\resizebox{\linewidth}{!}{
\begin{tabular}{llccccc}
\toprule
\rowcolor{MTitleGray}
Method & Setting & Art. F1 & Charge F1 & Sent. & 4-Step & Full-Chain \\
\midrule
DeepSeek-V3.2 & Zero-Shot & 0.924 & 0.908 & 0.512 & 0.958 & 0.771 \\
DeepSeek-V3.2 & ICL (3-shot) & 0.923 & 0.911 & 0.510 & 0.961 & 0.773 \\
GPT-5.2 & Zero-Shot & 0.912 & 0.894 & 0.489 & 0.947 & 0.744 \\
GPT-5.2 & ICL (3-shot) & 0.916 & 0.899 & 0.485 & 0.952 & 0.747 \\
\midrule
\bgB{Qwen2.5-3B} & \bgB{JPO} & \bgB{0.929} & \bgB{0.921} & \bgB{0.536} & \bgB{0.967} & \bgB{0.791} \\
\bgB{Qwen2.5-3B} & \bgB{JPO + ICL (3-shot)} & \bgB{\textbf{0.930}} & \bgB{\textbf{0.921}} & \bgB{\textbf{0.538}} & \bgB{\textbf{0.968}} & \bgB{\textbf{0.793}} \\
\bottomrule
\end{tabular}}
\vspace{-6pt}
\caption{In-context learning on JPO-Dataset, extending Table~\ref{tab:proprietary_zero_shot}.}
\vspace{-10pt}
\label{tab:icl}
\end{table}

\begin{table}[H]
\centering
\apptabstyle
\resizebox{\linewidth}{!}{
\begin{tabular}{lccc}
\toprule
\rowcolor{MTitleGray}
System & Params & Inference cost / 1k cases & One-time training \\
\midrule
DeepSeek-V3.2 (API) & -- & $\approx$ \$1.9 & -- \\
GPT-5.2 (API) & -- & $\approx$ \$9.7 & -- \\
\bgB{JPO Qwen2.5-3B (local)} & \bgB{3B} & \bgB{$\approx$ \textbf{\$0.16}} & \bgB{53 GPU-h} \\
\bottomrule
\end{tabular}}
\vspace{-6pt}
\caption{Cost comparison per 1,000 JPO-Dataset cases. API entries use public list pricing; the local entry amortizes A800 GPU time over inference.}
\vspace{-10pt}
\label{tab:cost}
\end{table}

\subsection{Teacher-Model Scaling}
\label{sec:appendix_teacher_scaling}

Appendix~\ref{appendix:different_teacher_model} verifies that JPO is robust to substituting DeepSeek-V2 for the default teacher. Here we extend that check across five teachers spanning different scales and providers, holding the student (Qwen2.5-7B-Instruct) and the entire RL configuration fixed.

Two patterns emerge from Table~\ref{tab:teacher_scaling}. First, JPO is robust to the teacher: regardless of which system generates the Stage-I rationales, the RL stage lifts SFT substantially and the final JPO models fall in a narrow band, spanning 0.930--0.941 article F1, 0.919--0.930 charge F1, and 0.532--0.558 sentence score. Even a 7B teacher yields a strong final model, so JPO does not depend on a single powerful teacher. Second, a stronger teacher mainly raises the SFT starting point, with the RL stage still adding effective gains on top; the more legally aligned Claude Opus 4.8 gives the best SFT start and a marginally higher JPO endpoint, which helps at the margin without changing the conclusion.

This also supplies internal evidence for the choice of a large teacher. As the teacher scales from 7B to 32B to 72B, the SFT starting point on sentence prediction rises monotonically from 0.401 to 0.412 to 0.422, so larger teachers do transfer better legal-reasoning trajectories to the student. This is consistent with systematic Chinese-legal evaluations that report a positive relationship between model scale and legal-reasoning ability, including the syllogism-oriented LAiW benchmark \citep{dai2025laiw} and the goal-directed-argument evaluation of CourtReasoner \citep{han2025courtreasoner}. The role of the teacher is to provide an initial structured reasoning trajectory rather than to upper-bound final performance, and after JPO's RL stage the student may approach or surpass the teacher on this specific task.

\begin{table}[H]
\centering
\apptabstyle
\resizebox{\linewidth}{!}{
\begin{tabular}{llcccc}
\toprule
\rowcolor{MTitleGray}
Teacher & Scale & SFT Sent. & JPO Art. & JPO Charge & JPO Sent. \\
\midrule
Qwen2.5-7B-Instruct & 7B & 0.401 & 0.930 & 0.919 & 0.532 \\
Qwen2.5-32B-Instruct & 32B & 0.412 & 0.934 & 0.923 & 0.544 \\
\bgB{Qwen2.5-72B-Instruct (default)} & \bgB{72B} & \bgB{0.422} & \bgB{0.937} & \bgB{0.928} & \bgB{0.551} \\
DeepSeek-V2 & MoE & 0.438 & 0.939 & 0.926 & 0.554 \\
Claude Opus 4.8 & prop. & 0.449 & \textbf{0.941} & \textbf{0.930} & \textbf{0.558} \\
\bottomrule
\end{tabular}}
\vspace{-6pt}
\caption{Teacher-model scaling with a fixed student (Qwen2.5-7B-Instruct) on JPO-Dataset. ``SFT Sent.'' is the sentence score of the Stage-I checkpoint before RL.}
\vspace{-10pt}
\label{tab:teacher_scaling}
\end{table}

\section{Prompt Template for Structured SFT}
\label{sec:appendix_prompts}

Figure~\ref{fig:sft_prompt_template} shows the prompt template used to elicit four-stage legal rationales from the teacher model during structured SFT.

\definecolor{PromptTitleBg}{RGB}{226,220,230}   
\definecolor{PromptBodyBg}{RGB}{248,246,243}    
\definecolor{PromptStepA}{RGB}{239,233,228}     
\definecolor{PromptStepB}{RGB}{232,236,238}     
\definecolor{PromptLine}{RGB}{185,178,186}      
\definecolor{PromptAccent}{RGB}{91,80,110}      

\begin{figure*}[t]
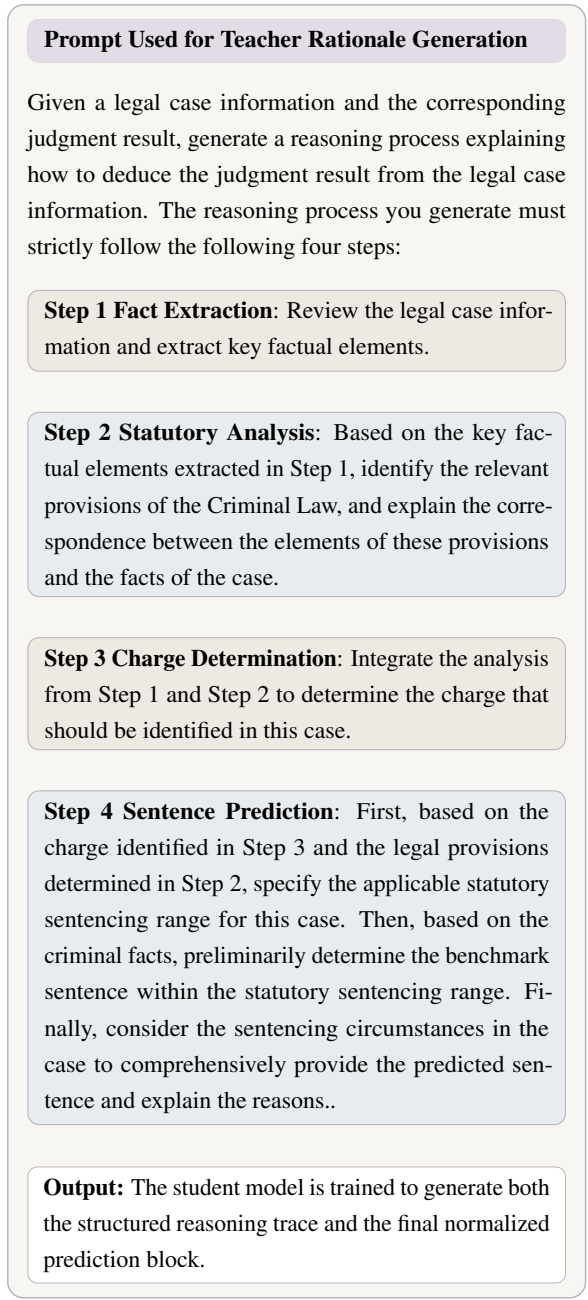

    \centering
    \begin{tcolorbox}[
        enhanced,
        width=\textwidth,
        colback=white,
        colframe=PromptLine,
        boxrule=0.55pt,
        arc=1.6mm,
        boxsep=0pt,
        left=1.3mm,
        right=1.3mm,
        top=1mm,
        bottom=1mm,
    ]
    \begin{tcolorbox}[
        enhanced,
        colback=PromptTitleBg,
        colframe=PromptTitleBg,
        boxrule=0pt,
        arc=1.1mm,
        left=1mm,
        right=1mm,
        top=0.7mm,
        bottom=0.7mm,
        before skip=0pt,
        after skip=3pt,
    ]
    \centering\small\bfseries\textcolor{PromptAccent}{Prompt Used for Teacher Rationale Generation}
    \end{tcolorbox}

    {\small
    \textbf{Instruction.}
    Given a legal case information and the corresponding judgment result, generate a reasoning process explaining how to deduce the judgment result from the legal case information. The reasoning process you generate must strictly follow the following four steps:
    \par}

    \vspace{3pt}
    \begin{tcbitemize}[
        raster columns=2,
        raster equal height=rows,
        raster column skip=2mm,
        raster row skip=2mm,
        enhanced,
        colframe=PromptLine,
        boxrule=0.35pt,
        arc=1.1mm,
        left=1.2mm,right=1.2mm,top=1mm,bottom=1mm,
        fontupper=\small,
    ]
    \tcbitem[colback=PromptStepA]
    \textcolor{PromptAccent}{\textbf{Step 1}}\hspace{0.5em}\textbf{Fact Extraction}\par
    \smallskip
    Review the legal case information and extract key factual elements.

    \tcbitem[colback=PromptStepB]
    \textcolor{PromptAccent}{\textbf{Step 2}}\hspace{0.5em}\textbf{Statutory Analysis}\par
    \smallskip
    Based on the key factual elements extracted in Step 1, identify the relevant provisions of the Criminal Law, and explain the correspondence between the elements of these provisions and the facts of the case.

    \tcbitem[colback=PromptStepA]
    \textcolor{PromptAccent}{\textbf{Step 3}}\hspace{0.5em}\textbf{Charge Determination}\par
    \smallskip
    Integrate the analysis from Step 1 and Step 2 to determine the charge that should be identified in this case.

    \tcbitem[colback=PromptStepB]
    \textcolor{PromptAccent}{\textbf{Step 4}}\hspace{0.5em}\textbf{Sentence Prediction}\par
    \smallskip
    First, based on the charge identified in Step 3 and the legal provisions determined in Step 2, specify the applicable statutory sentencing range for this case. Then, based on the criminal facts, preliminarily determine the benchmark sentence within the statutory sentencing range. Finally, consider the sentencing circumstances in the case to comprehensively provide the predicted sentence and explain the reasons.
    \end{tcbitemize}

    \vspace{3pt}
    \begin{tcolorbox}[
        enhanced,
        colback=PromptBodyBg,
        colframe=PromptLine,
        boxrule=0.35pt,
        arc=1.1mm,
        left=1.2mm,right=1.2mm,top=0.8mm,bottom=0.8mm,
        before skip=0pt,after skip=0pt,
    ]
    {\small\textbf{Output.} The student model is trained to generate both the structured reasoning trace and the final normalized prediction block.\par}
    \end{tcolorbox}
    \end{tcolorbox}
    \vspace{-6pt}
    \caption{Prompt template used to generate four-stage legal rationales for structured supervised fine-tuning.}
    \vspace{-8pt}
    \label{fig:sft_prompt_template}
\end{figure*}

\clearpage

\begin{figure}[h!]
	\centering
	\footnotesize
	\begin{tikzpicture}
		\node[draw, fill=background!30, rounded corners, inner sep=0pt, text width=1\textwidth, align=center] (case1) {
			\begin{minipage}[t]{\dimexpr\textwidth-0.2cm\relax}
				\setlength{\linewidth}{\dimexpr\textwidth-0.2cm\relax}
				\fcolorbox{primary}{primary!20}{\parbox{\dimexpr\textwidth-2\fboxsep-2\fboxrule\relax}{
						\centering\textcolor{primary}{\large\textbf{Case 1}}
				}}
				
				\vspace{0.2cm}
				
				\begin{minipage}{\dimexpr\textwidth\relax}
					\textcolor{textcolor}{\textbf{Fact:}} The prosecution alleged that on a certain afternoon in XX month XXXX year, the defendant Li Mouyi and his wife Ni Mou, at their residence near No. XX, XX Community, XXX Village, XX Street, XX City, had a dispute with their neighbor Li Moujia and his wife Du Mou over the placement of drainage pipes for a village road construction project. During the altercation, the defendant Li Mouyi pushed the victim Li Moujia, causing him to fall and fracture his right leg. Forensic examination determined that the victim Li Moujia suffered minor injuries, with a post-injury sequela assessed as a Level 10 disability. \\
					\textcolor{textcolor}{\textbf{Ground Truth:}} \{\textquotedblleft Article\textquotedblright: [\textquotedblleft 234\textquotedblright], \textquotedblleft Charge\textquotedblright: [\textquotedblleft Intentional Injury\textquotedblright], \textquotedblleft Sentence\textquotedblright: 12\} \\[0.3cm]
					
					\fcolorbox{step1color}{step1color!10}{\parbox{\dimexpr\linewidth-2\fboxsep-2\fboxrule\relax}{
							\textcolor{step1color}{\textbf{[Step 1]}} 
							\begin{itemize}[leftmargin=*, nosep]
								\item \textbf{Time:} XX month XXXX year, afternoon
								\item \textbf{Location:} Near No. XX, XX Community, XXX Village, XX Street, XX City
								\item \textbf{Parties:} Defendant Li Mouyi and his wife Ni Mou; Victim Li Moujia and his wife Du Mou
								\item \textbf{Cause:} Dispute over the placement of drainage pipes for a village road construction project
								\item \textbf{Event:} Li Mouyi argued with Li Moujia and Du Mou, then pushed Li Moujia, causing him to fracture his right leg
								\item \textbf{Injury:} Forensic examination: minor injuries, Level 10 disability
							\end{itemize}
					}}
					
					\vspace{0.2cm}
					
					\fcolorbox{step2color}{step2color!10}{\parbox{\dimexpr\linewidth-2\fboxsep-2\fboxrule\relax}{
							\textcolor{step2color}{\textbf{[Step 2]}}
							\begin{itemize}[leftmargin=*, nosep]
								\item \textbf{Legal Provision:} Article 234 of the Criminal Law of the People's Republic of China
								\item \textbf{Elements:}
								\begin{itemize}[leftmargin=*, nosep]
									\item \textbf{Intentional injury to another's body:} In this case, Li Mouyi intentionally pushed Li Moujia, causing him to fracture his right leg, which meets the element of "intentional injury to another's body."
									\item \textbf{Causing minor injury:} Forensic examination determined that Li Moujia suffered minor injuries, meeting the element of "causing minor injury."
								\end{itemize}
							\end{itemize}
					}}
					
					\vspace{0.2cm}
					
					\fcolorbox{step3color}{step3color!10}{\parbox{\dimexpr\linewidth-2\fboxsep-2\fboxrule\relax}{
							\textcolor{step3color}{\textbf{[Step 3]}}
							\begin{itemize}[leftmargin=*, nosep]
								\item \textbf{Charge:} Intentional injury
							\end{itemize}
					}}
					
					\vspace{0.2cm}
					
					\fcolorbox{step4color}{step4color!10}{\parbox{\dimexpr\linewidth-2\fboxsep-2\fboxrule\relax}{
							\textcolor{step4color}{\textbf{[Step 4]}}
							\begin{itemize}[leftmargin=*, nosep]
								\item \textbf{Statutory sentencing range:} According to Article 234 of the Criminal Law, intentional injury causing minor injury is punishable by up to three years imprisonment, criminal detention, or public surveillance.
								\item \textbf{Baseline sentence:} In this case, Li Moujia suffered minor injuries with Level 10 disability, which is considered a moderate degree of minor injury. The initial baseline sentence is estimated at around one year.
								\item \textbf{Sentencing circumstances:} The defendant Li Mouyi voluntarily surrendered and truthfully confessed his crime after returning to the case, constituting voluntary surrender. According to Article 67 of the Criminal Law, voluntary surrender may lead to a lighter or mitigated punishment.
								\item \textbf{Predicted sentence:} Considering the above factors, Li Mouyi's statutory sentence is up to three years. With the voluntary surrender, a lighter punishment is possible. The final predicted sentence is 10 months.
							\end{itemize}
					}}
					
					\vspace{0.2cm}
					
					\fcolorbox{answercolor}{answercolor!10}{\parbox{\dimexpr\linewidth-2\fboxsep-2\fboxrule\relax}{
							\textcolor{answercolor}{\textbf{[Answer]}} \{\textquotedblleft Article\textquotedblright: [\textquotedblleft 234\textquotedblright], \textquotedblleft Charge\textquotedblright: [\textquotedblleft Intentional Injury\textquotedblright], \textquotedblleft Sentence\textquotedblright: 10\} \\[0.3cm]
							\textcolor{textcolor}{\textbf{Metadata:}} $Score\_article=1$, $Score\_charge=1$, $Score\_sentence=\exp(-3 \times 1/6)\approx 0.61$
					}}
				\end{minipage}
				
				\vspace{0.3cm}
			\end{minipage}
		};
		
	\end{tikzpicture}
\end{figure}

\clearpage

\begin{figure}[h!]
	\centering
	\footnotesize
	\begin{tikzpicture}
		\node[draw, fill=background!30, rounded corners, inner sep=0pt, text width=1.0\textwidth, align=center, below=of case1, yshift=-0.5cm] (case2) {
			\begin{minipage}[t]{\dimexpr\textwidth-0.2cm\relax}
				\setlength{\linewidth}{\dimexpr\textwidth-0.2cm\relax}
				\fcolorbox{primary}{primary!20}{\parbox{\dimexpr\textwidth-2\fboxsep-2\fboxrule\relax}{
						\centering\textcolor{primary}{\large\textbf{Case 2}}
				}}
				
				\vspace{0.2cm}
				
				\begin{minipage}{\dimexpr\textwidth\relax}
					\textcolor{textcolor}{\textbf{Fact:}} It was established through investigation that at a certain hour on XX day of XX month XXXX year, the defendant Liu went to cut grass in the hills surrounding XXX Village. Upon seeing the victim Mo cutting down two small trees for firewood, which were tea seedlings from his responsibility mountain forest, he argued with the victim. Liu then used the back of his sickle blade to strike Mo's left forehead. Mo suffered a comminuted skull fracture and cerebral contusion (Grade 1 minor injury), an orbital fracture (Grade 2 minor injury), and a laceration on the left eyebrow (minor injury). Liu's family reached a criminal settlement agreement with Mo. Liu compensated Mo for medical expenses, lost wages, nursing fees and disability compensation. Mo expressed forgiveness for Liu's actions. \\
					\textcolor{textcolor}{\textbf{Ground Truth:}} \{\textquotedblleft Article\textquotedblright: [\textquotedblleft 234\textquotedblright], \textquotedblleft Charge\textquotedblright: [\textquotedblleft Intentional Injury\textquotedblright], \textquotedblleft Sentence\textquotedblright: 5\} \\
					
					\fcolorbox{step1color}{step1color!10}{\parbox{\dimexpr\linewidth-2\fboxsep-2\fboxrule\relax}{
							\textcolor{step1color}{\textbf{[Step 1]}}
							\begin{itemize}[leftmargin=*, nosep]
								\item \textbf{Time and Location:} XX hour on XX day of XX month XXXX year, surrounding hills of XXX Village
								\item \textbf{Perpetrator:} Defendant Liu
								\item \textbf{Victim:} Victim Mo
								\item \textbf{Specific act:} Liu used the back of his sickle blade and strike Mo's left forehead.
								\item \textbf{Injury result:} Forensic examination: comminuted skull fracture (Grade 1 minor injury), cerebral contusion (Grade 1 minor injury), orbital fracture (Grade 2 minor injury), laceration on the left eyebrow (minor injury)
								\item \textbf{Settlement:} Liu's family reached a criminal settlement agreement with Mo, compensating 22,000 yuan, and Mo expressed forgiveness
							\end{itemize}
					}}
					
					\vspace{0.2cm}
					
					\fcolorbox{step2color}{step2color!10}{\parbox{\dimexpr\linewidth-2\fboxsep-2\fboxrule\relax}{
							\textcolor{step2color}{\textbf{[Step 2]}}
							\begin{itemize}[leftmargin=*, nosep]
								\item \textbf{Legal Provision:} Article 234 of the Criminal Law
								\begin{itemize}[leftmargin=*, nosep]
									\item \textbf{Paragraph 1:} Intentional injury to another's body is punishable by up to three years imprisonment, criminal detention, or public surveillance.
									\item \textbf{Paragraph 2:} Whoever commits the crime in the preceding paragraph and causes serious injury shall be sentenced to fixed-term imprisonment of not less than three years but not more than ten years.
								\end{itemize}
							\end{itemize}
					}}
					
					\vspace{0.2cm}
					
					\fcolorbox{step3color}{step3color!10}{\parbox{\dimexpr\linewidth-2\fboxsep-2\fboxrule\relax}{
							\textcolor{step3color}{\textbf{[Step 3]}}
							\begin{itemize}[leftmargin=*, nosep]
								\item \textbf{Charge:} According to Article 234 of the Criminal Law, Liu Moumou's actions constitute the crime of intentional injury.
							\end{itemize}
					}}
					
					\vspace{0.2cm}
					
					\fcolorbox{step4color}{step4color!10}{\parbox{\dimexpr\linewidth-2\fboxsep-2\fboxrule\relax}{
							\textcolor{step4color}{\textbf{[Step 4]}}
							\begin{enumerate}[leftmargin=*, nosep]
								\item \textbf{Statutory sentencing range:}
								According to Article 234, intentional injury to another's body is punishable by up to three years imprisonment.
								\item \textbf{Baseline sentence:}
								 In this case, Mo's injuries were Grade 1 and Grade 2 minor injuries, within the range of up to three years imprisonment.
								
								\item \textbf{Sentencing circumstances:}
								\begin{itemize}[leftmargin=*, nosep]
									\item \textbf{Mitigating factors:}
									Liu's family reached a criminal settlement agreement with Mo, compensating 22,000 yuan, and Mo expressed forgiveness.
									
									\item \textbf{Aggravating factors:} No significant aggravating factors.
								\end{itemize}
								\item \textbf{Comprehensive assessment:}
								Considering Liu's actions caused Grade 1 and Grade 2 minor injuries to Mo, and his active compensation and receipt of forgiveness after the incident, a lighter punishment within the statutory range is possible.
									\item \textbf{Predicted sentence:} 8 months
								
							\end{enumerate}
					}}
					
					\vspace{0.2cm}
					
					\fcolorbox{answercolor}{answercolor!10}{\parbox{\dimexpr\linewidth-2\fboxsep-2\fboxrule\relax}{
							\textcolor{answercolor}{\textbf{[Answer]}} \{\textquotedblleft Article\textquotedblright: [\textquotedblleft 234\textquotedblright], \textquotedblleft Charge\textquotedblright: [\textquotedblleft Intentional Injury\textquotedblright], \textquotedblleft Sentence\textquotedblright: 8\} \\[0.3cm]
							\textcolor{textcolor}{\textbf{Metadata:}} $Score\_article=1$, $Score\_charge=1$, $Score\_sentence=\exp(-3 \times 3/5)\approx 0.17$
					}}
				\end{minipage}
				
				\vspace{0.1cm}
			\end{minipage}
		};
		
	\end{tikzpicture}
\end{figure}

\end{document}